%% file: anonymous-submission-latex-2023.tex
\documentclass[letterpaper]{article} % DO NOT CHANGE THIS
\usepackage[submission]{aaai23}  % DO NOT CHANGE THIS
\usepackage{times}  % DO NOT CHANGE THIS
\usepackage{helvet}  % DO NOT CHANGE THIS
\usepackage{courier}  % DO NOT CHANGE THIS
\usepackage[hyphens]{url}  % DO NOT CHANGE THIS
\usepackage{graphicx} % DO NOT CHANGE THIS
\usepackage{natbib}  % DO NOT CHANGE THIS AND DO NOT ADD ANY OPTIONS TO IT
\usepackage{caption} % DO NOT CHANGE THIS AND DO NOT ADD ANY OPTIONS TO IT
\usepackage{algorithm}
\usepackage{algorithmic}
\usepackage{cite}
\usepackage{amsmath,amssymb,amsfonts}
\usepackage{graphicx}
\usepackage{textcomp}
\usepackage{xcolor}
\usepackage{multirow}
\usepackage{subcaption}
\usepackage{cleveref}

\usepackage{newfloat}
\usepackage{listings}
\DeclareCaptionStyle{ruled}{labelfont=normalfont,labelsep=colon,strut=off} % DO NOT CHANGE THIS
\floatstyle{ruled}
\newfloat{listing}{tb}{lst}{}
\floatname{listing}{Listing}
\title{Bootstrapping Multi-view Representations for Fake News Detection{}}
\author{
    Qichao Ying,\textsuperscript{\rm 1}
    Xiaoxiao Hu, \textsuperscript{\rm 1}
    Yangming Zhou, \textsuperscript{\rm 1}
    Zhenxing Qian, \textsuperscript{\rm 1}
    Dan Zeng \textsuperscript{\rm 2}
    Shiming Ge \textsuperscript{\rm 3}
}
\affiliations{
   \textsuperscript{\rm 1} Fudan University, \textsuperscript{\rm 2} Shanghai University, \textsuperscript{\rm 3} Chinese Academy of Sciences\\
    \{qcying20,xxhu21\}@fudan.edu.cn, dzeng@shu.edu.cn, geshiming@iie.ac.cn
}

\usepackage{bibentry}
\begin{document}

\maketitle
\input{command}

\begin{abstract}
Previous researches on multimedia fake news detection include a series of complex feature extraction and fusion networks to gather useful information from the news. However, how cross-modal consistency relates to the fidelity of news and how features from different modalities affect the decision-making are still open questions. This paper presents a novel scheme of Bootstrapping Multi-view Representations (BMR) for fake news detection. Given a multi-modal news, we extract representations respectively from the views of the text, the image pattern and the image semantics. Improved Multi-gate Mixture-of-Expert networks (iMMoE) are proposed for feature refinement and fusion. Representations from each view are separately used to coarsely predict the fidelity of the whole news, and the multimodal representations are able to predict the cross-modal consistency. With the prediction scores, we reweigh each view of the representations and bootstrap them for fake news detection. Extensive experiments conducted on typical fake news detection datasets prove that the proposed BMR outperforms state-of-the-art schemes.

\end{abstract}

\section{Introduction}

Online Social Networks (OSNs) such as Weibo and Twitter have become important means of socializing and knowledge sharing.
The flourishing of OSN also leads to the fast spread of fake news and distorted opinions.
Fake news are specified as the news that are intentionally fabricated and can be verified as false~\cite{ruchansky2017csi,shu2017fake}.
In many cases, it is difficult for ordinary people to identify fake news. 
For network administrators, manually removing false news one by one is laborious and expensive.
Therefore, automatic Fake News Detection (FND) has become a hot research topic~\cite{allein2021like,shu2020fakenewsnet,xue2021detecting}. The FND techniques can effectively analyze the probability of misconducting information, which is more convenient for the administrator to block fake news from spreading on the OSN. 

The general paradigm of machine-learning-based fake news detection is to transform the news into a multidimensional latent representation, and identify the fidelity of the news using binary classification. 
Existing methods can be classified into three categories, namely, unimodal fake news detection~\cite{TM,shu2020leveraging}, multimodal fake news detection~\cite{MVAE,xue2021detecting} and dynamic fake news detection, i.e, utilizing propagation graph~\cite{qian2018neural} or knowledge graph~\cite{abdelnabi2022open,sun2022ddgcn}.
Some unimodal fake news detection methods can achieve good performances. However, most posts in OSN are in the multimodal style. Hence, the detection based on unimodal features is far from enough, and accordingly, we focus on multimodal fake news detection in this paper.

\begin{figure*}[!t]
  \centering
  \includegraphics[width=1.0\linewidth]{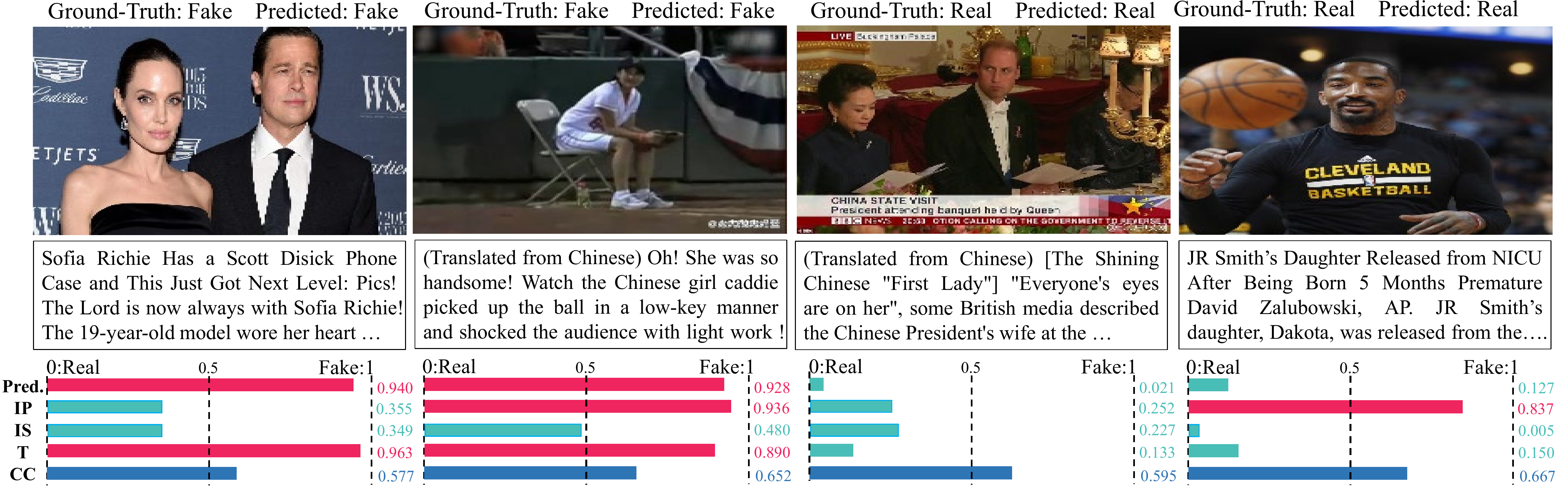}
  \caption{Examples of fake news detection result of BMR on Weibo (left two) and GossipCop (right two). The Chinese sentences from Weibo are translated here for references. Row ``IP" (Image Pattern), ``IS" (Image Semantics), ``T" (Text) respectively show the single-view prediction results provided by BMR, which suggests the dubious parts of the news. Row ``CC" (Cross-modal Consistency) shows the predictions on cross-modal consistency and Row ``Pred." provides the ultimate FND results.}
  \label{demo_results}
\end{figure*}

Many existing multimodal FND schemes use the textual and visual features as integrated representations~\cite{MVAE,SpotFake,chen2019attention,allein2021like}. 
Nevertheless, the disentanglement of features from different views has not been thoroughly investigated. In many cases, the models are at a black-box level, in which the network designs cannot explicitly highlight the most contributive components~\cite{EANN,SpotFake}.
Besides, many recent works solely rely on cross-modal correlation to generate fused features~\cite{wei2022cross,WWW}, but we argue that cross-modal correlation not necessarily play a critical role. 
Fig.~\ref{demo_results} provides four examples respectively from the well-known English GossipCop ~\cite{SpotFake} and the Chinese Weibo ~\cite{weibo} dataset, where both fake and real news more or less contain cross-modal correlations. 
Therefore, clarifying the roles of both unimodal and cross-modal features is vital for improving FND.

Aiming at addressing these issues, we propose an effective scheme that Bootstraps Multi-view Representations (BMR) for fake news detection.
Given a multi-modal news, we extract representations respectively from the views of the text, the image pattern and the image semantics. 
Improved Multi-gate Mixture-of-Expert networks (MMoE)~\cite{MMoE}, denoted as iMMoE, are proposed for feature refinement and fusion. 
Representations from each view are separately used to coarsely predict the fidelity of the whole news,
% and the cross-modal representations are used to predict the cross-modal consistency.
% After independently constructing single-view representations to coarsely predict the fidelity of the news, we generate multimodal representations to predict the cross-modal consistency. 
where the prediction scores are used for adaptive  feature reweighing.
Cross-modal consistency learning further implicitly guides multimodal representation refinement, and we explicitly disentangle correlation from other cross-modal information by introducing an independent representation.
Finally, we bootstrap multi-view features for refined fake news detection.
In the examples shown in Fig.~\ref{demo_results}, BMR not only predicts the fidelity of the news, but also gives confidence scores based on each view, which provides a new way of understanding how different roles act in multimodal FND. 

The contributions of this paper are three-folded:

\begin{itemize}
\item We propose a novel fake news detection scheme of generating multi-view representations, understanding their individual importance, and optimizing the fused features.

\item We propose to disentangle information within unimodal and multimodal features by single-view prediction and cross-modal consistency learning, which are then adaptively reweighed and bootstrapped for better detection.

\item The proposed BMR detection not only outperforms state-of-the-art multimodal FND schemes on popular datasets, but also provides a mechanism for interpreting the contributions of different representations.
\end{itemize}
 
% Our code is attached in the supplementary materials.

\section{Related Works}
\noindent\textbf{Unimodal Fake News Detection.}
Both the language-based and the vision-based fake news detections have received extensive attention. 
% The linguistic modality is relatively stable, in that most news and posts on the OSN contain text descriptions, while other modalities such as images or videos are kind of sporadic.
% Generally, language-based fake news detection methods first convert the text into feature vectors, and then use the network with attention mechanism for feature extraction.
% Experiments show that the improved attention mechanism leads to significant performance improvement in fake news detection.
% There are also some works that jointly judge the degree of falsehood for all news under an event.
% % The news belonging to the same event is listed together and sent to the network for joint feature extraction and analysis. 
% The motivation of this design is to extract the common and distinctive features from the posts under the same event. 
TM~\cite{TM} utilizes lexical and semantic properties of the text to detect fake news.
MWSS~\cite{shu2020leveraging} exploits multiple weak signals from different sources from user and content engagements.
% LSTM-ATT~\cite{lin2019detecting} build traditional known machine learning models to extract 134 features and uses a LSTM with self-attention mechanism to see which model performs better.
% Other works take into consideration user response to the targeted news~\cite{allein2021like,qian2018neural}.
% Feng et al.~\cite{qian2018neural} finds that how people response to historical articles can be learnt by a generative model, and the generated user comments can provide evidence about why articles are judged as fake news. They accordingly propose a novel two-stage convolutional neural network with user response generator.
% , where TCNN captures semantic information from article text by extracting information at the sentence and word level. URG obtains article text and corresponding real comments from historical user responses.
% \noindent\textbf{Visual Fake News Detection.}
% Aside from language-based fake news detection schemes, researchers also distinguish fake news with reference to their images. 
Jin et al.~\cite{jin2016novel} find that there are noticeable differences in the distribution of images between real news and fake news.
Cao et al.~\cite{Leveraging_5} suggest that typical methods for image manipulation detection~\cite{MVSS} are useful in unveiling traces for news tampering.
% Also, fake news publishers often use images to attract and mislead readers in order to spread rapidly. 
% For instance, the number of photos of fake news tends to be much simpler than that of real news, because many fake news tend to use the same news as photos.
Besides, Qi et al.~\cite{qi2019exploiting} 
% As a result, fake news images tend to show visual impact and emotional provocation. 
jointly use the spatial domain and frequency domain features of the news for forensics.
% These methods have good performances. 
However, for the multimodal news, these approaches are unable to detect cross-modal correlations.
% Besides, image manipulation detecion schemes~\cite{MVSS} are also useful in unveiling traces of news tampering.

\begin{figure*}[!t]
  \centering
  \includegraphics[width=1.0\linewidth]{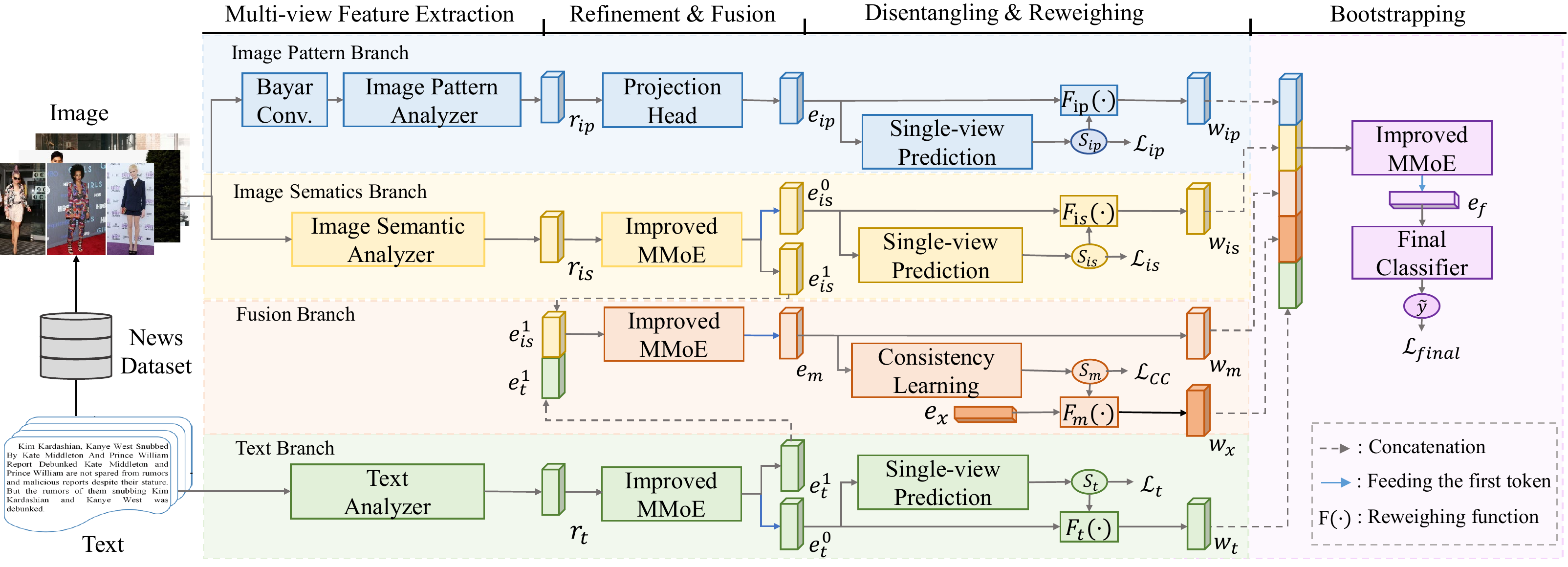}
  \caption{The network architecture of BMR. Representations are respectively extracted from the views of text, image pattern and image semantics. Improved MMoE, together with single-view predictions and cross-modal consistency learning, jointly guides unimodal feature refinement and cross-modal feature generation. The final decision is made upon bootstrapping these multi-view representations. 
  }
%   We mine multi-view representations from the news to make coarse predictions. Cross-modal consistency learning aids the generation of multimodal representations. All features are adaptively reweighed and bootstrapped to make more accurate prediction result.
  \label{image_architecture}
\end{figure*}

\noindent\textbf{Multimodal Fake News Detection.}
The critical issue of multimodal fake news detection is to align linguistic and vision representations.
SpotFake~\cite{SpotFake} integrates pretrained XLNet and ResNet for feature extraction.
% The unimodal features are then concatenated by multiple fully connected layers to conduct the final classification.
% Chen et al.~\cite{chen2021multimodal} introduces a latent event memory module to store event semantic information.
SAFE~\cite{zhou2020mathsf} feeds the relevance between news textual and visual information into a classifier to detect fake news.
% LIIMR~\cite{singhal2022leveraging} extracts relevant information from the strong modality on a per-sample basis.
EANN~\cite{EANN} introduces an additional discriminator to classify news events so as to suppress the impact of specific events on classification.
% Qi et al.~\cite{Entity-Enhanced} effectively captures multimodal cues by extracting visual entities as well as text within the image to grasp more information from the news.
MCAN~\cite{wu2021multimodal} stacks multiple co-attention layers for multimodal feature fusion.
% MVAE~\cite{MVAE} uses a variational autoencoder for multimedia to compress the news into the probabilistic latent variables, and the classifier predicts whether the news is real or fake based on the latent variables.
% Aiming at the situation that images and texts are irrelevant in news, Sun et al.~\cite{sun2021inconsistency} proposes a knowledge-guided two-stream inconsistency network based on the intuition that fake news is more likely to contain inconsistent information semantically.
% The two networks respectively extract cross-modal differences, exclude the shared information and identify anomalous entity pairs co-occurring in news content by measuring their indicative distances.
% Li et al.~\cite{Entity-Enhanced} proposes an entity-oriented multimodal alignment and fusion network (EMAF), which adoptes entity-centric cross-modal interaction and designed an alignment module based on an improved dynamic routing algorithm.
MCNN~\cite{xue2021detecting} also incorporates textual semantic features, visual tampering features, and similarity of textual and visual information for fake news detection, but the aggregation process only concatenates all features. Hence, how each of these multi-view features affects the predictions cannot be measured.
In comparison, we explicitly reweigh multi-view representations based on single-view classifications and adaptively bootstrap them for fake news detection.

% \noindent\textbf{Dynamic Fake News Detection.}
% Besides, some posts might contain novel or little knowledge which cannot find a match in KG or online searching. 
% Therefore, this paper only conduct fake news detection using the static information within the news. 
% \noindent\textbf{Fake News Detection with Cross-modal Correlation.}

Besides, some methods propose to use cross-modal correlation learning for fake news detection.
% EMAF~\cite{Entity-Enhanced} adopts entity-centric cross-modal interaction and designed an alignment module based on an improved dynamic routing algorithm.
CAFE~\cite{WWW} uses a VAE to compress images and texts representations and measures cross-modal consistency based on their Kullback-Leibler (KL) divergence. The consistency score then linearly adjusts the weight of unimodal and multimodal features before final classification.  
Likewise, CMC~\cite{wei2022cross} includes a two-staged network that train two uni-modal networks to learn cross-modal correlation by contrastive learning, and then finetune the network for fake news detection.
However, CAFE severely penalizes unimodal features when the consistency is high, which might be doubtful in some news. CMC does not adaptively suppress or augment multi-view representations, and the fine-tuning stage may erase the cross-modal knowledge learned in the first stage. 
Therefore, we propose an improved mechanism to better utilize cross-modal consistency learning for FND.

% Attention~\cite{vaswani2017attention,wang2018non} allows a model to learn to make predictions by selectively attending to a given set of data.
% Attention-based learning algorithms has become the first choice for many natural language processing and computer vision tasks. 
% The transformer network is originally proposed by Vaswani et al.~\cite{vaswani2017attention}, where the main components are multi-head self-attention blocks and feed-forward networks. The design of transformer stems from the classic sequence-to-sequence RNN model, which can be split into two parts, namely, encoder and decoder.
% Since then, the transformer architecture has undergone various application studies and model enhancements. 
% For example, 
% % Sparce Transformer~\cite{child2019generating} eases the computational complexity by adopting local attention and strided attention.
% Linformer~\cite{wang2020linformer} reduces the length dimension of key and value matrices with additional projection layers.
% Performer~\cite{choromanski2020rethinking} leverages orthogonal random features to approximate the attention matrix with provable accuracy.
% The emerging of the transformer architecture~\cite{vaswani2017attention,he2022masked,BERT} has given rise to many Vision-Language-Pretraining (VLP) models, such as VilBERT~\cite{vilbert}, OSCAR~\cite{OSCAR}, ViLT~\cite{vilbert}. These models are trained on large corpus of several millions of text-image pairs to align the representation between vision and language.

\section{Proposed Approach}
% \subsection{Pipeline Design}
%综合讲一讲4 branches, 4 phases
% Analyzer Phase
% Refine Phase
% Classification & Reweighing Phase
% Bootstrapping Phase
Fig.~\ref{image_architecture} depicts the pipeline of the proposed Bootstrapping Multi-view Representations approach, which contains four stages, i.e., Multi-view Feature Extraction, Refine \& Fusion, Disentangling \& Reweighing,  and Bootstrapping. In the first three stages, there are four branches corresponding to the representations from four views, including the Image Pattern Branch, Image Semantics Branch, Text Branch, and Fusion Branch.

\subsection{Multi-view Feature Extraction}
Let the input multimodal news be $\mathcal{N}=[\mathbf{I},\mathbf{T}]\in\mathcal{D}$, where $\mathbf{I},\mathbf{T},\mathcal{D}$ are the image, the text and the dataset, respectively.
We begin with extracting coarse multi-view representations, including the image pattern, the image semantics, and the text, corresponding to the Image Pattern Branch, the Image Semantic Branch, and the Text Branch, respectively. We denote these representations as ${r}_{\emph{ip}}$, ${r}_{\emph{is}}$ and ${r}_{\emph{t}}$. 

We believe that the general distribution of the image and the tiny traces left by tampering or compression can be helpful to expose fake news. Therefore, we explicitly separate the feature learning paradigm from image pattern and semantics.
% , which is different from many previous methods that use a single network to learn image features~\cite{EANN,SpotFake}. 
In the Image Pattern Branch, we use InceptionNet-V3~\cite{inceptionv3} as the image pattern analyzer. 
We additionally include a BayarConv~\cite{bayar2018constrained} to process $\mathbf{I}$ ahead of sending it into the InceptionNet-V3 network, since BayarConv augments details and suppresses the main component of $\mathbf{I}$ that contributes largely to semantics. 
While CNNs are suitable for vision learning, the transformer models~\cite{he2022masked} based on masked language/image modeling are reported to be great in establishing long-range attention~\cite{ViT}.
Hence, in the Image Semantic Branch, we use Masked Autoencoder (MAE) as the image semantics analyzer to extract ${r}_{\emph{is}}$.  
In the Text Branch, we use BERT~\cite{BERT} to extract ${r}_{\emph{t}}$.
% Besides, image higher-frequent details empirically share little information with image semantics and text representations. Therefore, we regard image pattern and image semantics as two individual modalities and disentangle the visual information mining task using two networks.

\begin{figure}[!t]
  \centering
  \includegraphics[width=0.9\linewidth]{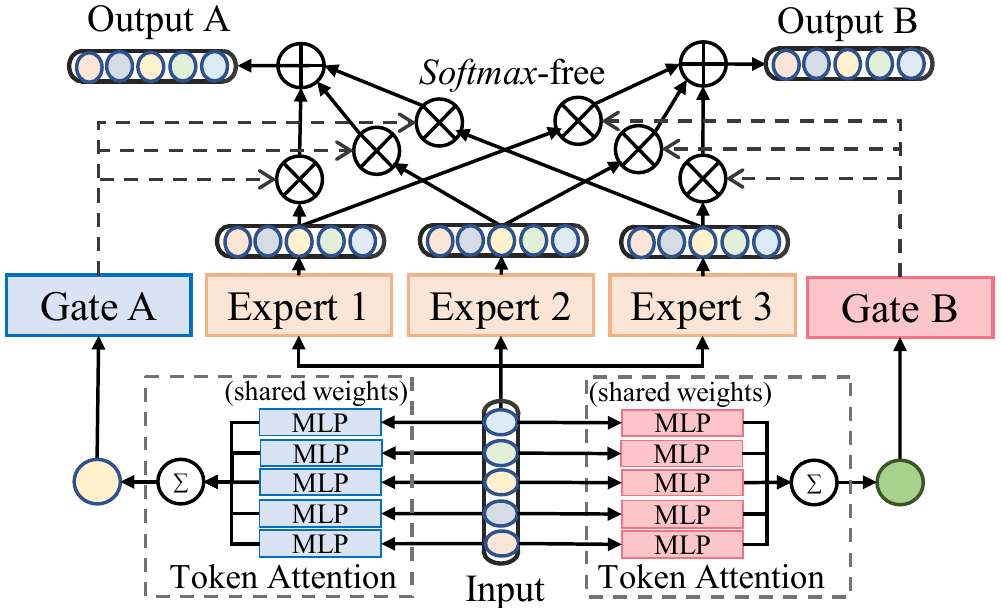}
  \caption{Network architecture of improved MMoE.}
  \label{image_mmoe}
\end{figure}
\subsection{Feature Refinement and Fusion with iMMoE}

In the Refine \& Fusion Stage, we refine the representations extracted from the first stage. 
In the Image Pattern Branch, 
the representation ${r}_{\emph{ip}}$ is projected to a new representation ${e}_{\emph{ip}}$, where the classification head of the InceptionNet-V3 is replaced with an MLP-based projection head. The output size of ${e}_{\emph{ip}}$ equals to that of a single token representation from MAE. 
In the Image Semantics Branch and the Text Branch, we propose the improved MMoE (iMMoE) network to refine ${r}_{\emph{is}}$ and ${r}_{\emph{t}}$. We also generate a new representation ${e}_{\emph{m}}$ in the Fusion Branch.

As sketched in Fig.~\ref{image_mmoe}, the proposed iMMoE contains several experts, gates and token attentions that produces features in separate branches. 
The MMoE network~\cite{MMoE} is originally designed to model multi-task relationships from data by sharing the expert across all tasks. The input $x$ is equally sent into $n$ expert networks, and $k$ gates adaptively weigh the outputs of the experts as the final output using $\emph{softmax}$ function. $n$ is a hyper-parameter and $k$ is the amount of down-stream tasks. The algorithm of MMoE is summarized in Eq.~(\ref{eqn_original})
\begin{equation}
x^{k}=\sum_{i=1}^{n} \emph{softmax}(G^{k}_{i}(x))\cdot~E_{i}(x),
\label{eqn_original}
\end{equation}
where $E_{i}$ and $G^{k}_{i}$ are the $i_{\emph{th}}$ expert and the $i_{\emph{th}}$ output of the gate for task $k$.
In BMR, we treat the mining of multi-view unimodal representations and cross-modal feature fusion as different subtasks. They should share common features and reserve their own distinctive features. 
% However, directly applying MMoE on token representations is computationally expensive.
We improve MMoE network in two aspects. First, we use token attention to compute the importance scores of each token representation using an MLP, and perform dimension reduction by aggregating all token representations into one according to the scores. The aggregated representation is then sent into the gate to compute the weights for each expert. 
Second, we find that the $\emph{softmax}$ functions in the gates are used to guarantee that the output weights are all positive and the sum equals one. This is not necessary according to our experiments. Accordingly, we lift the $\emph{softmax}$ constraints and allow the weights to be outside the $[0,1]$ range.  
In summary, we revise Eq.~(\ref{eqn_original}) as Eq.~(\ref{eqn_gate}) for iMMoE,
% Besides, no $\emph{softmax}$ is used in calculating the token attention.
\begin{equation}
x^{k}=\sum_{i=1}^{n}(G_{i}^{k}({\sum_{j=1}^{t}\emph{MLP}_{k}(x)})\cdot~E_{i}(x)),
\label{eqn_gate}
\end{equation}
where $t$ is the amount of tokens and $\emph{MLP}_{k}$ denotes the token attention for task $k$. $k\in[1,2]$. 

With the iMMoE network, we refine ${r}_{\emph{is}}$ and ${r}_{\emph{t}}$ into the features $[{e}_{\emph{is}}^0,{e}_{\emph{is}}^1]$ and $[{e}_{\emph{t}}^0,{e}_{\emph{t}}^1]$, respectively. ${e}_{\emph{is}}^0$ and ${e}_{\emph{t}}^0$ are preserved for single-view prediction. Meanwhile, the ${e}_{\emph{is}}^1$ and ${e}_{\emph{t}}^1$ are jointly fed into the iMMoE net in the Fusion Branch to generate a multimodal feature ${e}_{\emph{m}}$, which is preserved for cross-modal consistency learning and bootstrapping. 

\subsection{Disentangling \& Reweighing}
After the refinement and fusion, we obtain the new representations from four branches. Next, we handle representations of image pattern, image semantics and text via single-view prediction, while processing the fused representation in the Fusion Branch for consistency learning.

\noindent\textbf{Single-view Prediction and Reweighing}.
% The disentangling \& reweighing stage is composed of four MLP-based classifiers respectively for single-view prediction and cross-modal consistency learning. 
Single-view prediction uses the first token of ${e}_{\emph{is}}^0$, the first token of ${e}_{\emph{t}}^0$  and ${e}_{\emph{ip}}$, to predict the fidelity of $\mathcal{N}$. 
We design this module based on two considerations. 
On one hand, many fake news contains obvious abnormality in either image noise, distribution or text independently. Therefore, making predictions on $\mathcal{N}$ with single-view features is empirically feasible that allows the network to deeply mine the characteristics from each view.
On the other hand, the prediction scores can be projected into weights that adaptively reweigh the representations.
Therefore, we use MLP to project each single-view representation into scores, and use another MLP, i.e., $F(\cdot)$ in Fig.~\ref{image_architecture}, to project the scores into weights.
Take the generation of $S_{\emph{ip}}$ and $w_{\emph{ip}}$ as an example,
\begin{equation}
\begin{gathered}
S_{\emph{ip}}=\emph{MLP}_\emph{ip}(e_{\emph{ip}}),\\
w_\emph{ip}=\emph{Sigmoid}\left(\emph{F}_{ip}\left(S_{\emph{ip}}\right)\right)\cdot{e}_\emph{ip},
\end{gathered}
\end{equation}
where $\emph{MLP}_{\emph{ip}}(\cdot)$ denotes the MLP-based single-view predictor. In the same way, we generate paired prediction scores and reweighed representations $\{S_\emph{is},w_\emph{is}\}$, $\{S_m,w_\emph{m}\}$ and $\{S_\emph{t},w_\emph{t}\}$. 
The reweighed representations are then ready for bootstrapping.

\begin{figure}[!t]
  \centering
  \includegraphics[width=1.0\linewidth]{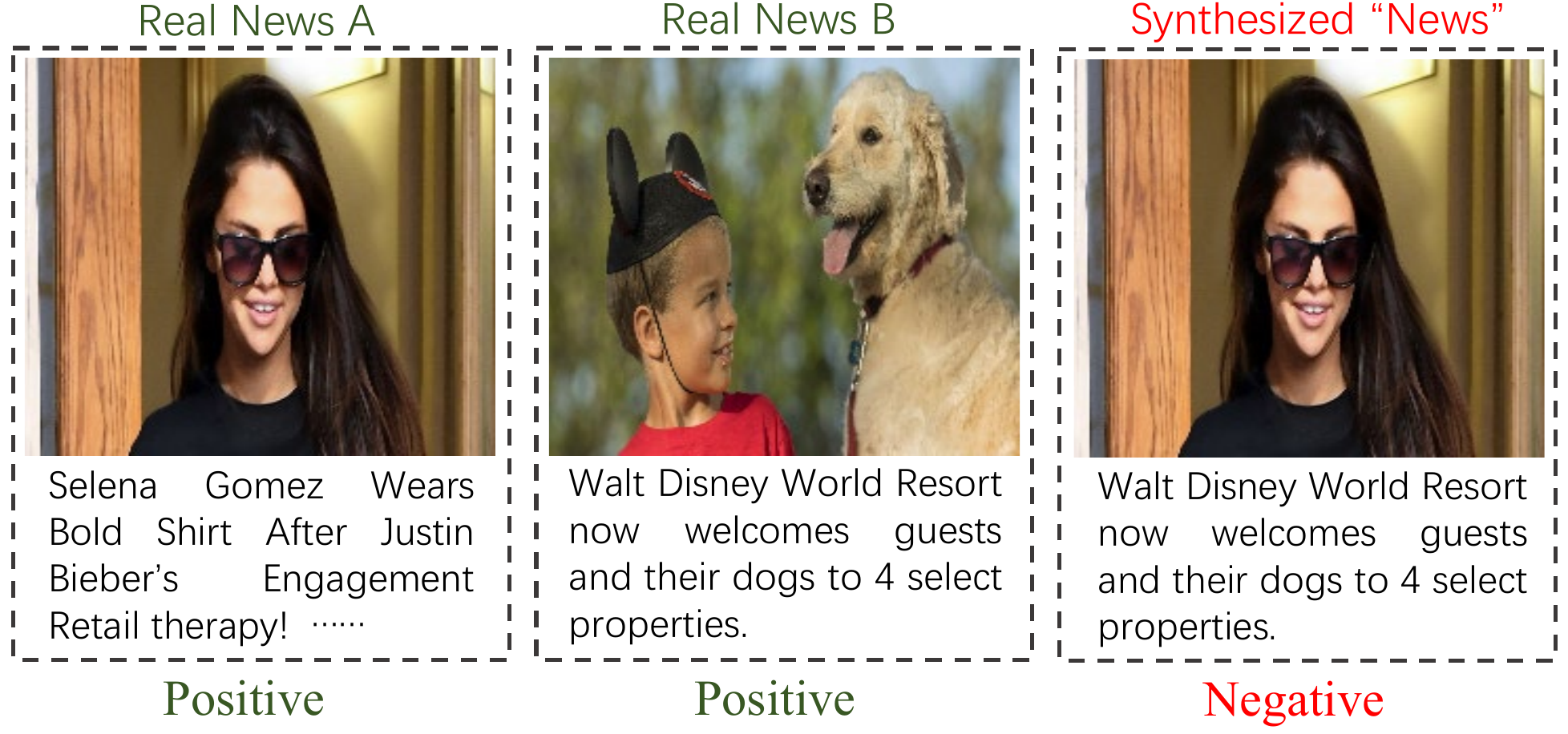}
  \caption{Training data for cross-modal consistency learning. Real news borrowed from typical FND datasets are marked positive and synthesized news by combining image and text from different real news are marked negative.}
  \label{image_cross_modal}
\end{figure}
% \begin{figure*}[!t]
%   \centering
%   \includegraphics[width=1.0\linewidth]{figure/demo.pdf}
%   \caption{Examples of fake news detection using BMR. Attentions scores which are over 0.8 are marked in blue. Three coarse classification might provide different results. After feature aggregation, the ultimate fake news detection result is more precise than the coarse ones.}
%   \label{experiment}
% \end{figure*}

% \subsection{Improved MMoE Network with Token Attention}

\noindent\textbf{Cross-modal Consistency Learning}. 
Inspired by several recent works~\cite{WWW,wei2022cross}, we guide the multimodal feature fusion in the Fusion Branch with cross-modal consistency learning. 
Specifically, we train BMR to predict whether a given text-image pair matches. We begin with crafting a new dataset $\mathcal{D}'=[\mathcal{D}_{\emph{real}},\mathcal{D}_{\emph{syn}}]$ on the basis of $\mathcal{D}$, where news with correlated texts and images are labeled $y'=1$, otherwise $y'=0$.
Fig.~\ref{image_cross_modal} shows three training data for cross-modal consistency learning, where the positive examples are directly the real news borrowed from $\mathcal{D}$ and the negative examples are synthesized ``news" by arbitrarily combining images and texts from different real news. 
%It contain equal number of positive and negative examples, whose sum is close to that of $\mathcal{D}$.
Though there can be cases that some news in $\mathcal{D}_{\emph{real}}$ do not contain cross-modal consistency, the possibility of text-image pairs from $\mathcal{D}_{\emph{syn}}$ having consistency is much lower in nature. 
% i.e., the false negative rate caused by dataset synthesis is naturally way lower than the false positive rate. 
Therefore, the model can still learn the correlation based on such a hybrid dataset.
We feed BMR with  $\mathcal{N}'=[\mathbf{I}',\mathbf{T}']\in\mathcal{D}'$. After the corresponding $e_{\emph{m}}$ is calculated, we use an MLP-based predictor to output the consistency score ${S}_{\emph{m}}$, and expect the score to be close to the label $y'$. 
We use a simple regression loss as supervision instead of contrastive loss applied in CMC~\cite{wei2022cross} in that the likelihood of any text from the negative pool matching the query image, i.e., the false negative rate, is proportional to the size of that pool.
The training process of cross-modal consistency learning only activates a part of the modules in BMR. The task can be in parallel learned with the main task.
% It is worth mentioning that in BMR, $e_{\emph{m}}$ not only contains information for predicting the consistency, it is also used for the bootstrapping module. 
% Also, though the example in Fig~\ref{demo_results} shows that cross-modal correlation has limited direct effect on the fidelity of the news, teaching BMR to predict if the image and text are correlated helps it better mine useful multimodal representations.

\noindent\textbf{Reweighing Multimodal Representation}. 
% We also reweigh multimodal representations according to the predicted score of cross-modal consistency score. 
The predicted consistency scores on $\mathcal{N}$ also adjust the multimodal representations.
We consider that multimodal features should not merely represent cross-modal correlation. There are also many other factors, e.g., joint distribution, emotional difference, etc., that can help decide whether the news is fake.
Therefore we refrain from weighing $e_\emph{m}$ using $S_m$.
Instead, we explicitly disentangle correlation from other cross-modal information by letting $S_m$ reweigh an independent trainable token $e_x$ that represents \textit{cross-modal irrelevance}. $w_x={e}_{\emph{x}}\cdot~F_{\emph{m}}(S_m)$ and $w_m=e_m$, which are both considered multimodal representations for bootstrapping.
Compared to previous works, we do not separate the cross-modal learning with the detecting stage~\cite{wei2022cross} or reweighing both unimodal and multimodal representations by correlation score~\cite{WWW}.
% We refrain from weighing $e_\emph{m}$ using $S_m$, 
% We allow $w_m$ to contain information supplemental to the multi-view unimodal representations.
% $w_{\emph{ip}}$, $w_{\emph{is}}$ and $w_{\emph{t}}$. 

% Therefore, we do not reweigh $e_{\emph{m}}$ using $S_{\emph{m}}$ but 

\begin{algorithm}[tb]
\caption{Model training of BMR}
\label{algorithm}
\textbf{Input}: Dataset: $\mathcal{D}$, Training epochs: $N$, Amount of news for constructing $\mathcal{D}'$: $k$\\
\textbf{Output}: Model parameters: $\Theta$.
\begin{algorithmic}[1] %[1] enables line numbers
\STATE Sample $k/2$ real news from $\mathcal{D}$ and store them into $\mathcal{D}'$ as positive examples.
\FOR{i in range($k$/4):}
\STATE Sample $\mathcal{N}_1=[\mathbf{I}_1, \mathbf{T}_1]$, $\mathcal{N}_2=[\mathbf{I}_2, \mathbf{T}_2]$ from $\mathcal{D}$.
\STATE Synthesize pseudo-news by shuffling information in $\mathcal{N}_1$ and $\mathcal{N}_2$. $\mathcal{N}_3=[\mathbf{I}_1, \mathbf{T}_2]$, $\mathcal{N}_4=[\mathbf{I}_2, \mathbf{T}_1]$.
\STATE Store $\mathcal{N}_3, \mathcal{N}_4$ into $\mathcal{D}'$ as negative examples.
\ENDFOR
\FOR{i in range(N):}
\STATE Sample $(\mathbf{I}, \mathbf{T},y)$, $(\mathbf{I}', \mathbf{T}', y')$ from $\mathcal{D}$, $\mathcal{D}'$ 
\STATE $S_{\emph{m}}=\emph{BMR}(\mathbf{I}', \mathbf{T}', $\emph{train\_consist}$=\emph{True})$
\STATE Compute loss using $\mathcal{L}_{\emph{BCE}}(y',  S_{\emph{m}})$.
\STATE $\mathbf{T}=$``\textit{No text provided.}" if $\emph{len}(\mathbf{T})<5$.
% \ENDIF
\STATE $\mathbf{I}=\emph{zeros\_like}(\mathbf{I})$ if $\mathbf{I}.\emph{rows}<64$ or $\mathbf{I}.\emph{cols}<64$
\STATE [$\tilde{y}, S_{\emph{t}}, S_{\emph{is}}, S_{\emph{ip}}]=\emph{BMR}(\mathbf{I}, \mathbf{T},  $\emph{train\_consist}$=\emph{True})$.
\STATE Compute loss using $\mathcal{L}_{\emph{BCE}}(y,\tilde{y})$, $\mathcal{L}_{\emph{BCE}}(y,S_{\emph{t}})$, $\mathcal{L}_{\emph{BCE}}(y,S_{\emph{is}})$, $\mathcal{L}_{\emph{BCE}}(y,S_{\emph{ip}})$.
\STATE Update parameters in $\Theta$ using Adam optimizer.
\ENDFOR
% \STATE \textbf{return} solution
\end{algorithmic}
\end{algorithm}
\subsection{Bootstrapping Stage and Loss Function} 
The multi-view representations $[{w}_{\emph{is}}, {w}_{\emph{ip}},{w}_{\emph{m}},{w}_{\emph{x}},{w}_{\emph{t}}]$ are bootstrapped using another iMMoE, which further refines the information critical for the decision-making. The final MLP-based classifier gets the first token of the output $e_f$ to predict $\tilde{y}$, which is expected to be close to the label $y$.

Fake news detection is a binary classification problem. 
The Binary Cross-Entropy loss (BCE) is defined in Eq.~(\ref{eqn_bce}), where $x$ and $\tilde{x}$ stand for the ground truth and the prediction, respectively.
\begin{equation}
\mathcal{L}_{\emph{BCE}}(x,\tilde{x})=x \log \left(\tilde{x}\right)+\left(1-x\right) \log \left(1-\tilde{x}\right).
\label{eqn_bce}
\end{equation}

We apply the BCE loss between the ground-truth label
$y$ and the predicted scores $\tilde{y}$, as well as the coarse classification results $S_{\emph{ip}}$, $S_{\emph{is}}$, $S_{t}$. There is an extra loss for cross-modal consistency training, where we also apply the BCE loss between the ground-truth crafted label $y'$ and $S_{\emph{m}}$.
Therefore, the losses are defined as $\mathcal{L}_{\emph{CC}}=\mathcal{L}_{\emph{BCE}}(y',S_{\emph{m}})$, $\mathcal{L}_{\emph{final}}=\mathcal{L}_{\emph{BCE}}(y,\tilde{y})$, $\mathcal{L}_{T}=\mathcal{L}_{\emph{BCE}}(y,S_{\emph{t}})$, $\mathcal{L}_{\emph{ip}}=\mathcal{L}_{\emph{BCE}}(y,S_{\emph{ip}})$ and $\mathcal{L}_{\emph{is}}=\mathcal{L}_{\emph{BCE}}(y,S_{\emph{is}})$.
The single view FND classification loss is the aggregate of $\mathcal{L}_{\emph{is}}$, $\mathcal{L}_{\emph{ip}}$ and $\mathcal{L}_{\emph{t}}$, namely,
\begin{equation}
\mathcal{L}_{\emph{coarse}}=(\mathcal{L}_{\emph{is}}+\mathcal{L}_{\emph{ip}}+\mathcal{L}_{\emph{t}})/3,
\end{equation}
The total loss for BMR is defined as follows.
\begin{equation}
\mathcal{L}=\mathcal{L}_{\emph{final}}+\alpha\cdot\mathcal{L}_{\emph{coarse}}+\beta\cdot\mathcal{L}_{\emph{CC}},
\label{eqn_total_loss}
\end{equation}
where the hyper-parameters $\alpha$ and $\beta$ are empirically set as $\alpha=1$, and $\beta=4$.
We provide the pseudo-code of detailed training processes of BMR in Algo.~\ref{algorithm}.

\begin{table*}[!t]
\centering
  \resizebox{1.0\textwidth}{!}{
\begin{tabular}{ccccccccc}
\hline
\multirow{2}{*}   & \multirow{2}{*}{Method} &\multirow{2}{*}{Accuracy} &\multicolumn{3}{c}{Fake News}  & \multicolumn{3}{c}{Real News}\\
   \cline{4-9}&       &     & Precision   & Recall  & F1-score    & Precision    & Recall      & F1-score             \\
\hline
\multirow{6}{*}{Weibo} 

    & EANN*~\cite{EANN}   & 0.827 & 0.847 & 0.812     & 0.829     & 0.807       & 0.843      & 0.825        \\
% & MVAE~\cite{MVAE}    & 0.824    & 0.854   & 0.769    & 0.809    & 0.802  & 0.875   & 0.837  \\
% & Spotfake~\cite{singhal2019spotfake}    & 0.892    & 0.902  & \textbf{0.964}   & \textbf{0.932}   & 0.847    & 0.656    & 0.739   \\
 & MCNN~\cite{xue2021detecting}     & 0.846       & 0.809   & 0.857 & 0.832     & 0.879      & 0.837      & 0.858    \\
% &SAFE~\cite{zhou2020mathsf}     &0.762   &0.831  &0.724  &0.774  &0.695  &0.811  &0.748 \\
% &LIIMR~\cite{singhal2022leveraging}    & \bluemarker{\textbf{0.900}}    & 0.882   & 0.823   & 0.847    & \bluemarker{\textbf{0.908}}   & \bluemarker{\textbf{0.941}}   & \redmarker{\textbf{0.925}}  \\
 & MCAN~\cite{wu2021multimodal}      & \bluemarker{\textbf{0.899}}   & \bluemarker{\textbf{0.913}}   & \bluemarker{\textbf{0.889}}   & \bluemarker{\textbf{0.901}}   & \bluemarker{\textbf{0.884}}   & 
 \bluemarker{\textbf{0.909}}   & 0.897   \\
& CAFE*~\cite{WWW}    & 0.840     & 0.855  & 0.830  & 0.842  & 0.825   & 0.851  & 0.837    \\
& CMC~\cite{wei2022cross}    & 0.893     & \redmarker{\textbf{0.940}}  & 0.869  & 0.899  & 0.876   & \redmarker{\textbf{0.945}}  & \redmarker{\textbf{0.907}}    \\
& \textbf{BMR~(Proposed)}   & \redmarker{\textbf{0.918}}  & 0.882 & \redmarker{\textbf{0.948}}  & \redmarker{\textbf{0.914}} & \redmarker{\textbf{0.942}} & 0.870  & \bluemarker{\textbf{0.904}} \\
\hline
% \multirow{7.5}{*}{Politifact} 
% & MWSS~\cite{shu2020leveraging}   & 0.820    &  & -     & -  & 0.820   & -     & -    \\
%   & SAFE~\cite{zhou2020mathsf}   & 0.874   & 0.851    & 0.830    &0.840   & 0.889   & 0.903   & 0.896   \\
%  & Spotfake+~\cite{SpotFake}  & 0.846  & -   & -    & -   & -  & -   & -    \\
%  & TM~\cite{TM}   & 0.871  & -    & -     & -   & 0.901 & -    & -   \\
% & LSTM-ATT~\cite{lin2019detecting}   & 0.832  & 0.828  &0.832   & 0.830  & 0.836  & 0.832 & 0.829  \\
% & DistilBert~\cite{allein2021like}  & 0.741    & 0.875 & 0.636  & 0.737 & 0.647 & 0.880  & 0.746 \\
% & CAFE~\cite{WWW}  & 0.864 & 0.724 & 0.778 & 0.750 & 0.895 & 0.919    & 0.907  \\
% & BMR & \textbf{0.942}   & \textbf{0.897} & \textbf{0.897}   & \textbf{0.897}    & \textbf{0.960} & \textbf{0.960}& \textbf{0.960}  \\
% \hline
\multirow{6}{*}{GossipCop} 
% & MWSS~\cite{shu2020leveraging} & 0.800  & -    & -   & 0.800   & - & -   & - \\
& EANN*~\cite{EANN}  & 0.864  & 0.702  & 0.518   & 0.594     & 0.887  & 0.956  & 0.920  \\
 & Spotfake*~\cite{SpotFake}  & 0.858   & 0.732 & 0.372 & 0.494    & 0.866    & 0.962    & 0.914  \\
% & TM~\cite{TM}  & 0.842 & - & -   & - & 0.896   & -  & -  \\
% & LSTM-ATT~\cite{lin2019detecting}  & 0.842   & -   & -   & -  & 0.839  & 0.842 & 0.821  \\
& DistilBert~\cite{allein2021like}  & 0.857  & \bluemarker{\textbf{0.805}}  & 0.527  & 0.637  & 0.866& 0.960 & 0.911\\
& CAFE*~\cite{WWW}  & 0.867 & 0.732  & 0.490  & 0.587   & 0.887& 0.957  & 0.921 \\
& CMC~\cite{wei2022cross} & \bluemarker{\textbf{0.893}}     & \redmarker{\textbf{0.826}}  & \redmarker{\textbf{0.657}}  & \redmarker{\textbf{0.692}}  & \redmarker{\textbf{0.920}}   & \bluemarker{\textbf{0.963}}  & \bluemarker{\textbf{0.935}}    \\
& \textbf{BMR~(Proposed)} & \redmarker{\textbf{0.895}}  & 0.752 & \bluemarker{\textbf{0.639}}  & \bluemarker{\textbf{0.691}}       & \redmarker{\textbf{0.920}}    & \redmarker{\textbf{0.965}} & \redmarker{\textbf{0.936}}    \\
    \hline
    \multirow{4}{*}{Weibo-21} 
% & MWSS~\cite{shu2020leveraging} & 0.800  & -    & -   & 0.800   & - & -   & - \\
& EANN*~\cite{zhou2020mathsf}  & 0.870  & 0.902  & 0.825   & 0.862     & 0.841  & \redmarker{\textbf{0.912}}  & 0.875  \\
& SpotFake*~\cite{WWW}  & 0.851 & \redmarker{\textbf{0.953}}  & 0.733  & 0.828   & 0.786 & 0.964  & 0.866 \\
& CAFE*~\cite{WWW}  & \bluemarker{\textbf{0.882}} & 0.857  & \bluemarker{\textbf{0.915}}  & \bluemarker{\textbf{0.885}}   & \bluemarker{\textbf{0.907}} & 0.844  & \bluemarker{\textbf{0.876}} \\
& \textbf{BMR~(Proposed)} & \redmarker{\textbf{0.929}}  & \bluemarker{\textbf{0.908}} & \redmarker{\textbf{0.947}}  & \redmarker{\textbf{0.927}}       & \redmarker{\textbf{0.946}}    & \bluemarker{\textbf{0.906}} & \redmarker{\textbf{0.925}}    \\      
    \hline
  \end{tabular}}
    \caption{Comparison between BMR and state-of-the-art multimodal fake news detection schemes on Weibo, GossipCop and Weibo-21. *: open-sources. The best performance is highlighted in bold red and the follow-up is highlighted in bold blue.}
  \label{comparison}
\end{table*}

\section{Experiments}
We have conducted many experiments using popular datasets to verify the proposed BMR approach for fake news detection. The source codes along with more results are attached as the supplementary material.

\subsection{Experimental Setups}
\noindent\textbf{Implementation Details.}
We use the ``mae-pretrain-vit-base" model for image processing, the ``bert-base-chinese” model for Chinese dataset, and the ``bert-base-uncased” model for English dataset. The hidden sizes of both MAE and BERT are $768$.
BERT and MAE are kept frozen.
All MLPs in BMR contain one hidden layer, a BatchNorm1D and an ELU activation.
We use the single-layer transformer blocks defined in ViT~\cite{ViT} to implement the experts in iMMoE.
% with sinusoidal position encoding.
Each iMMoE network holds three experts.
We use Adam optimizer with the default parameters. The batch size is 24 and the learning rate is $1\times10^{-4}$ with cosine annealing decay.
Images are resized into $224\times224$, and the max length of text is set as $197$.
BMR is also designed to be compatible with unimodal news by feeding a \textit{zero matrix} as the image or \textit{``No text provided"} as the text. Similar to EANN~\cite{EANN}, we increase the quality of the datasets by replacing images smaller than $64\times64$ and texts less than five words.

\noindent\textbf{Data Preparation.}
We use Weibo~\cite{weibo}, GossipCop~\cite{shu2020fakenewsnet} and Weibo-21~\cite{MDFEND} for training and testing. 
Weibo contains 3749 real news and 3783 fake news for training, 1000 fake news and 996 real news for testing. 
GossipCop contains 7974 real news and 2036 fake news for training, 2285 real news and 545 fake news for testing.
Weibo-21 is a newly-released dataset that contains 4640 real news and 4487 fake news in total, and we split it into training and testing data at a ratio of $9:1$. 
Though MediaEval~\cite{Twitter} and Politifact~\cite{SpotFake} are also popular datasets, they contain only 460 images and 381 posts in the training set. Hence, and neural nets training with Politifact can easily overfit with such small amount of data. 
We train BMR on each dataset five times with different initial weights and report the averaged best performance.

According to the theory in \cite{imbalance}, we determine a fixed threshold for each dataset based on its distribution. Since the ratio of real news versus fake news in GossipCop training set is close to 4:1, we set the threshold for GossipCop as 0.80. Similarly, the thresholds for Weibo and Weibo-21 are set as 0.50. 
For real-world applications, we can set a default threshold as 0.50.
% Since the ratio of real news versus fake news in GossipCop training set is close to 4:1, we accordingly over-sample the minority group, i.e., fake news, three times during training. We do not resample the test set.

% F1 score reports the overall precision and sensitivity of the scheme. The definition is as follows. 
% \begin{equation}
%     \emph{F1}=\frac{2\emph{TP}}{2\emph{TP}+\emph{FN}+\emph{FP}},
% \end{equation}
% where $\emph{TP}$ stands for True Positive rate, $\emph{FN}$ stands for False Negative rate, and $\emph{FP}$ stands for False Positive rate. Higher F1 value indicates more accurate classification result.

% \begin{figure}[!t]
%   \centering
%   \includegraphics[width=1.0\linewidth]{figure/bar.pdf}
%   \caption{Bar of Weibo.}
%   \label{image_bar}
% \end{figure}

\begin{figure}[!t]
  \centering
  \includegraphics[width=1.0\linewidth]{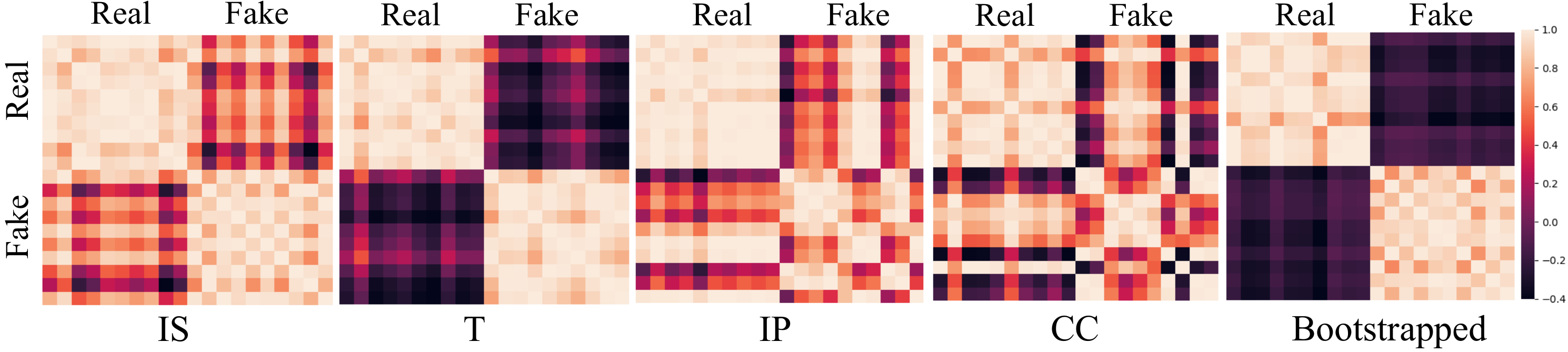}
  
  \caption{Heatmap Visualization. Each cell in the heat maps represents the paired cosine similarity between the 64-dim representations respectively from the final classifier and the single-view predictors. Tests are done on Weibo.}
%   Visualization of the discriminative capability of the multi-view representations.
  \label{image_heatmap}
\end{figure}

\subsection{Performance Analysis}
\noindent\textbf{Heatmap Visualization.}
In Fig.~\ref{image_heatmap}, we measure the discriminative capability of the multi-view representations using heatmap visualization. We arbitrarily select ten real news and ten fake news, and then compute paired similarity between the 64-dim representations from the final classifier and the single-view predictors. 
% Each sub-figure therefore shows the correlation pattern between inter-class news and intra-class news provided with different views of representations. 
From the figures, the bootstrapped representation shows strong discriminative capability where intra-class similarity and inter-class difference are both noticeable. Some other views of representations also have decent borders, but the heatmap for cross-modal consistency tends to be much messier, showing that it is not practical to directly use the consistency for FND.

\noindent\textbf{Comparisons.}
Table~\ref{comparison} shows the average precision, recall, and accuracy of BMR on Weibo and GossipCop. 
We use Accuracy, Precision, Recall, and F1 score for comprehensive performance measurements.
The results are promising, with 91.8\% average accuracy on Weibo, 92.9\% on Weibo-21 and 89.5\% on GossipCop. 
We further compare BMR with state-of-the-art methods. 
BMR outperforms the other methods on the datasets. Besides, we rank either $1\emph{st}$ or $2\emph{nd}$ on Recall and F1 score of fake news on all of the datasets.
CMC has a very close performance with BMR on GossipCop. However, on Weibo, BMR outperforms CMC by a more noticeable 2.5\%.
We have also trained three open-sourced methods on Weibo-21 and compared them with BMR. The result shows that BMR leads by 4.2\% in overall accuracy and 3.2\% in Recall of fake news, which proves the effectiveness of BMR.

Moreover, we present the t-SNE visualizations of features in Fig. 6,  which are learned by BMR, SpotFake and CAFE on the test set of GossipCop. In BMR, the dots representing fake news are comparatively farther from differently-labeled dots, and there are fewer outliers. This indicates that the extracted features in BMR are more discriminative.

%------ refrained from moving to supplement-------
\noindent\textbf{Computational Complexity.}
It costs around 12 hours to train BMR for 50 epochs on a single NVIDIA RTX 3090 GPU.
The computational complexity study in Table~\ref{table_complexity} shows that BMR requires less trainable parameters compared to EANN and SpotFake. 
CAFE requires amazingly low computational complexity by using simple auto-encoders and MLPs, but the performance is not good enough.
% Besides, BMR requires less trainable parameters than many multimodal pretraining models such as OSCAR~\cite{OSCAR}.

\begin{figure}[!t]
  \centering
  \includegraphics[width=0.92\linewidth]{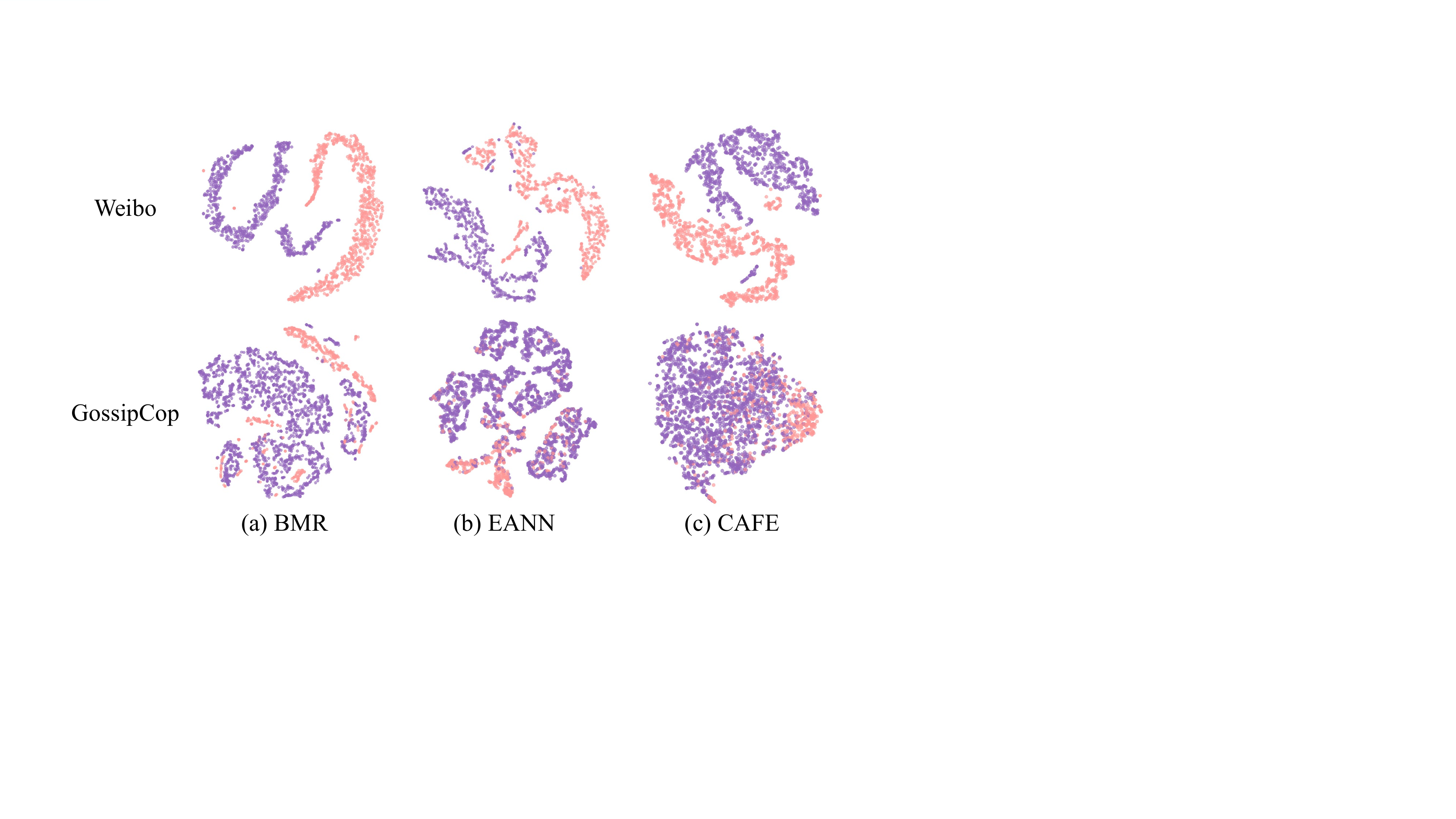}
  \caption{TSNE visualization of mined features on the test set. Dots with the same color are within the same label. }
  \label{image_tsne}
\end{figure}

\begin{table}[!t]
	\begin{center}
	\resizebox{0.42\textwidth}{!}{\begin{tabular}{c|c|c|c|c}
		\hline
	 & BMR & EANN  & SpotFake & CAFE   \\
        \hline
        Params & 94.39M & 143.70M & 124.37M & 0.68M \\
        FLOPS & 18.42G & 19.63G & 30.42G & 0.01G  \\
		\hline
	\end{tabular}}
	\caption{Comparison of trainable parameters and computational speed. FLOPs: amount of floating point arithmetics.}
	\label{table_complexity}
	\end{center}
\end{table}

\subsection{Ablation Study}
\label{section_ablation}
% We conduct ablation tests to investigate the effectiveness of each component in BMR. 

\begin{figure}[!t]
  \centering
  \includegraphics[width=1.0\linewidth]{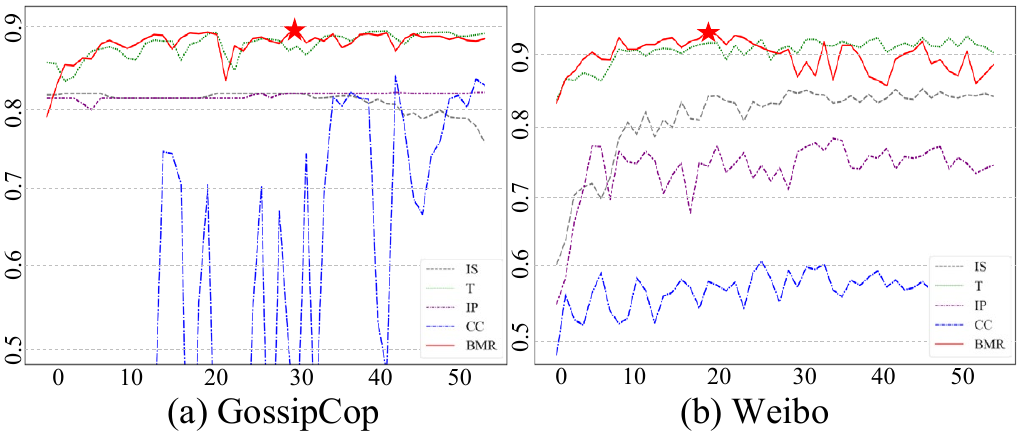}
  \caption{Curves of testing accuracy (y-axis) versus training epochs (x-axis). ``CC" curve represents using cross-modal consistency as the judgement of fake news detection.}
%   We see little trace of overfitting and high correlation between coarse text classification and the bootstrapped result.
  \label{image_overfit}
\end{figure}

\noindent\textbf{Contribution of Each View.} 
% The purpose of training coarse classifiers is to reweight different views of the representations according to the prediction. 
Fig.~\ref{image_overfit} and Table.~\ref{table_training_loss} show the curves for testing accuracy and ultimate training loss. 
We study the contribution of each view on the datasets.
% Given fake news, sometimes the classifiers are forced to predict fake though there is no faulty in some modalities.
First, we find that $\mathcal{L}_{\emph{ip}}$, $\mathcal{L}_{\emph{is}}$ and $\mathcal{L}_{\emph{CC}}$ do not fully converge on GossipCop. Therefore we closer scrutinized GossipCop dataset and surprisingly find that close to 50\% images are celebrity faces that are even hard to distinguish for many human viewers.
However, BMR still defeats all compared state-of-the-art FND schemes in the overall accuracy,
% and we find a strong correlation between the overall testing curve with the single-view prediction from text. 
which indicates that FND on GossipCop relies severely on text and images might contribute marginally.
Second, we study whether there is under-fitting, i.e., single-view predictions struggling with the ground truth $y$, or over-fitting, i.e., networks remembering the training set.
In Fig.~\ref{image_overfit}, the testing accuracies remain high and the curves go into plateau rather than drop even though we train 50 epochs. Therefore, there is no overfitting in our training.

\begin{table}[!t]
	\begin{center}
	\resizebox{0.43\textwidth}{!}{\begin{tabular}{c|c|c|c|c|c}
		\hline
	 Dataset & $\mathcal{L}_{\emph{final}}$ & $\mathcal{L}_{\emph{CC}}$  & $\mathcal{L}_{\emph{t}}$ & $\mathcal{L}_{\emph{is}}$ & $\mathcal{L}_{\emph{ip}}$  \\
        \hline
        Weibo & 0.005 & 0.084 & 0.007 & 0.026 & 0.133  \\
        GossipCop & 0.005 & 0.540 & 0.006 & 0.511 & 0.493 \\
        Weibo21 & 0.002 & 0.124 & 0.003 & 0.067 & 0.012 \\
		\hline
	\end{tabular}}
	\caption{Averaged ultimate training loss of each term.}
	\label{table_training_loss}
	\end{center}
\end{table}

%%%%%%% move to supplement

% We also surprisingly find that the accuracy based on cross-modal consistency starts to increase at the end of training.
% We thus suspect that in later epochs, the model starts to recognize some of these celebrities.

\noindent\textbf{Removal of Some Views.} 
In the upper part of Table~\ref{table_ablation}, we feed BMR with different combinations of views of representations. For example, in row three, we only feed BMR with the image, preserve the Image Pattern Branch and the Image Semantics Branch, while disabling the rest of the modules.
We find that if we bootstrap three multi-view unimodal representations without introducing the fusion branch, the averaged accuracy is 88.6\%, which is 3.2\% worse than BMR.  
Though we find that the contribution of text is large for FND, merely using texts only guarantee 87.0\% accuracy.
% Text representations seem to be the most powerful one. 
% From Fig.~\ref{image_overfit}, we also find that the fidelity of text shows high correlation with the bootstrapped result on GossipCop.
% Note that the result of coarse text classification is different from training BMR with text only, because the text representation is further reweighted and used by the bootstrapping module.

\begin{table}[!t]
\centering
  \resizebox{0.48\textwidth}{!}{
\begin{tabular}{cccc}
\hline
\multirow{2}{*}{Test} &\multirow{2}{*}{Accuracy} &\multicolumn{2}{c}{F1 Score} \\
  \cline{3-4}  &   & Fake News & Real News \\

\hline
IP  & 0.789  & 0.774   & 0.806   \\
IS  & 0.838  & 0.860   & 0.809   \\
IS+IP  & 0.857  & 0.874  & 0.833   \\
T  & 0.870  & 0.872  & 0.868   \\
IP+T  & 0.878  & 0.876  & 0.876   \\
IS+T  & 0.881  & 0.888  & 0.873   \\
IS+IP+T  & 0.886  & 0.888  & 0.884   \\
\hline
${S}_\emph{m}$ reweigh. Multi-view. & 0.864  & 0.854 & 0.809\\
w/o Feature Reweigh. & 0.887  & 0.880 & 0.892\\
w/o Coarse Class. & 0.895 & 0.903 & 0.887\\
using ViT Blocks for Refine. & 0.905  & 0.910 & 0.900\\
w/o Cross. Correlat. & 0.905  & 0.906 & 0.902\\
using ViT instead of MAE & 0.910  & 0.911  & 0.900\\
w/o improving MMoE & 0.907  & 0.910  & 0.899\\
\textbf{BMR}  & \textbf{0.918}  & \textbf{0.914} & \textbf{0.904} \\
\hline
\end{tabular}}
    \caption{Ablation study on bootstrapping multi-view representations and network design. The tests are done on Weibo. Results of GossipCop and Weibo-21 are in the supplement.}
\label{table_ablation}
\end{table}

\noindent\textbf{Cross-modal Consistency Learning. }
% Provided with all four multi-view representations, we further test the necessity of cross-modal consistency learning, coarse classifications and the introduction of MMoE network.
In Table~\ref{table_ablation}, we mimic CAFE that reweighs $e_{\emph{ip}},e_{\emph{is}},e_{\emph{t}}$ using ${S}_\emph{m}$. We surprisingly find that the result is even worse than not using multimodal features. It suggests that granting priority to cross-modal consistency rather than unimodal representations is not beneficial. Second, we remove cross-modal consistency learning and therefore add no constraint on multimodal feature generation. The performance drops 1.3\% accordingly.
% , proving that learning cross-modal consistency indeed helps increasing the network's cognitive function. 
From Table~\ref{table_training_loss} and Fig.~\ref{image_overfit}, we also find that though the training losses for consistency learning on Weibo and Weibo-21 are both low, we fail to conduct FND simply based on cross-modal consistency, where the accuracy is less than 0.6. Since we have ruled out over-fitting, we conclude that the proposed training methodology for cross-modal consistency is feasible but the consistency can so far only play an assistant rather than a critical role in multimodal FND.

\noindent\textbf{Single-view Prediction. }
In Table~\ref{table_ablation}, we further test the case of adding no constraint to the generation of multi-view representation, which results in a performance that drops 2.3\%. Therefore, single-view prediction is a useful manual bias compared to blindly employing multi-branches for feature extraction.
Besides, if we do not explicitly reweigh the representations according to the sub-tasks, the accuracy will decrease by 2.9\%, indicating that there would be misleading information if we do not reweigh the representations. 

\begin{figure}[!t]
  \centering
  \includegraphics[width=1.0\linewidth]{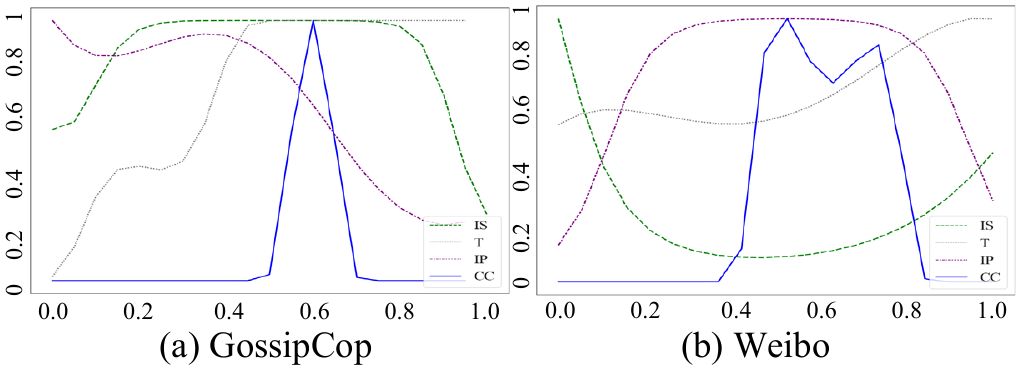}
  \caption{Curves of the learned reweighing functions $F(\cdot)$. x-axis: score, e.g., $S_{\emph{ip}}$, y-axis: corresponding weight. }
%   We see little trace of overfitting and high correlation between coarse text classification and the bootstrapped result.
  \label{image_score}
\end{figure}

Fig.~\ref{image_score} shows the learned reweighing functions $F(\cdot)$ on GossipCop and Weibo.
We expected ``V"-shaped functions to prefer single-view prediction score with more confidence, i.e., close to either zero or one. However, many of these learned functions exhibit a inverted ``V“ shape. 
Besides, $e_x$ is only activated when the cross-modal consistency score is close to 0.5 on Weibo.

\noindent\textbf{Network Design.}
Since we are the first to use MAE in FND, we verify the case of replacing MAE with ViT. The overall accuracy of BMR drops 0.8\% results in Table.~\ref{table_ablation}, which indicates that MAE is better for the task. Besides, When using iMMoE instead of separate ViT blocks, the result will increase by 1.3\%, suggesting that sharing some representations is beneficial.
We also find that the internal improvement made in MMoE in BMR helps gaining 1.1\%.

% \begin{figure}[!t]
%   \centering
%   \includegraphics[width=1.0\linewidth]{figure/demo_tamper.pdf}
%   \caption{Sensitivity test. We distort the meaning of the two real news in Fig.~\ref{demo_results}. The difference is marked red. The predictions are different from the original ones accordingly.}
%   \label{image_modification}
% \end{figure}

% \noindent\textbf{Sensitivity test.}
% In Fig.~\ref{image_modification}, we manually distort the two real news provided in Fig.~\ref{demo_results} where the meaning has been altered. We invited ten volunteers and they all agreed that the modified news should be considered fake. From the results, we see that the predictions correlated to the manipulated modality are different from the original ones in Fig.~\ref{demo_results}, suggesting that BMR is sensitive to these differences.

\section{Conclusions}
We propose BMR that generates multi-view representations, understands the individual importance, and optimizes the fused features.
Single-view prediction and cross-modal consistency learning are proposed to disentangle information within unimodal and multimodal features, which are then adaptively reweighed and bootstrapped for better detection results. Experiments show that BMR outperforms state-of-the-art multimodal FND schemes on popular datasets.

\bibliography{anonymous-submission-latex-2023}

\end{document}

\maketitle
\input{command}

\begin{figure}[!t]
  \centering
  \includegraphics[width=1.0\linewidth]{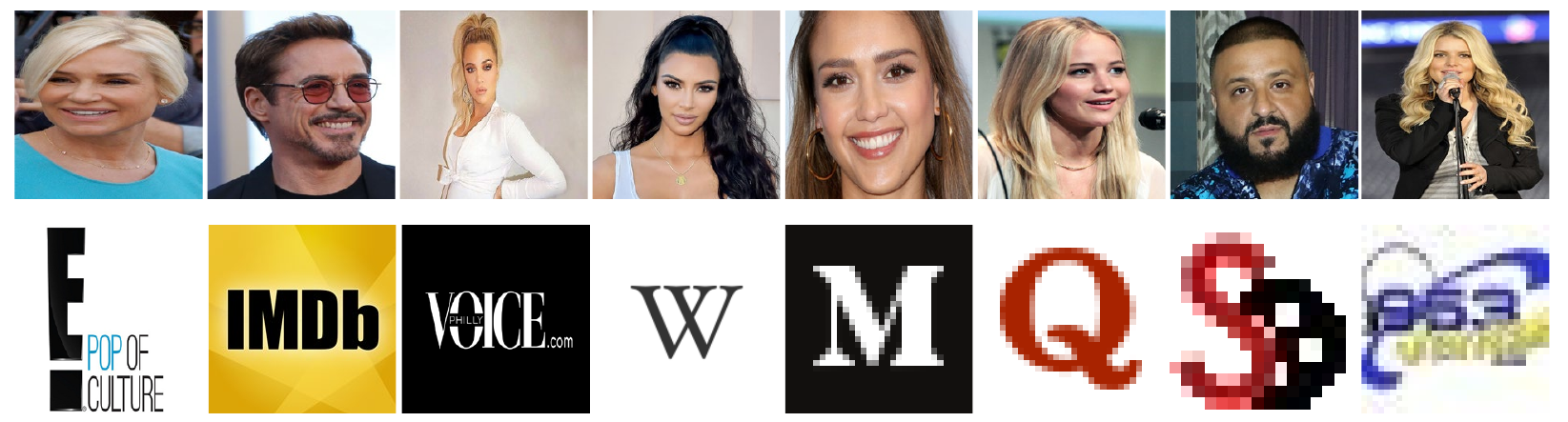}
  \caption{Image characteristics of GossipCop. We found a bunch of celebrity faces (upper row) and tiny-sized logos (lower row) both in the training and test set, which makes it hard for the networks to mine useful representations from the image modality.}
  \label{image_illegal}
\end{figure}
\section{More Implementation Details}
To prevent the reweighing process from affecting the single-view prediction stage as well as the cross-modal consistency learning stage, we block the backward gradient flow from $w$ to $S$. Take $S_{\emph{ip}}$ and $w_{\emph{ip}}$ as an example, Eq.(3) in the main paper can be more specifically defined as
\begin{equation}
\begin{gathered}
S_{\emph{ip}}=\emph{MLP}_\emph{ip}(\emph{Stop\_grad}(e_{\emph{ip}})),\\
w_\emph{ip}=\emph{Sigmoid}\left(\emph{F}_{ip}\left(S_{\emph{ip}}\right)\right)\cdot{e}_\emph{ip}.
\end{gathered}
\end{equation}

\section{Additional Experimental Results}
\subsection{Ablation Studies on GossipCop and Weibo-21}
We have also conducted ablation studies on GossipCop and Weibo-21, respectively, and from the results, we can draw similar conclusions to those on Weibo. 

\noindent\textbf{On GossipCop.}
The results are shown in Table.~\ref{table_ablation_gossip}.
Please note that real news in GossipCop accounts for 80.74\% of the total news on it test set, so the expectation of accuracy by random guessing is 80.74\%. As a result, merely using images as input on GossipCop cannot distinguish fake news from the real in that the accuracy is even worse than random guessing.
We surprisingly find that using text modality as input and BERT \& improved MMoE as feature extraction and refinement modules, the overall performance can be up to 88.3\%, which is only 1.2\% lower than the full implementation of BMR, and is higher than many previous state-of-the-art schemes such as EANN and CAFE. 
The results suggest that if the text representations can be well learned and exploited, images in GossipCop can only marginally contribute to the fake news detection results.
We showcase some images in Fig.~\ref{image_illegal} to further show the image characteristics of GossipCop. We find a bunch of celebrity faces (upper row) and tiny-sized logos (lower row) both in the training and test set, which makes it hard for the networks to mine useful representations from the image modality.
Also we find that if we use text and image pattern or semantics, the performance will be worse than using text only, suggesting that improper usage of image information can even hamper the overall performance. 

\noindent\textbf{On Weibo-21.}
The results are shown in Table.~\ref{table_ablation_weibo21}.
We find that using text only performs poorly where the testing accuracy is 48.3\%. Also, there is a wider gap between the performance of the full implementation of BMR and the performances of using only some of the multi-view representations. Once we use all the multi-view representations, the performance can be close to that of the full implementation.

\begin{table}[!t]
\centering
  \resizebox{0.48\textwidth}{!}{
\begin{tabular}{cccc}
\hline
\multirow{2}{*}{Test} &\multirow{2}{*}{Accuracy} &\multicolumn{2}{c}{F1 Score} \\
  \cline{3-4}  &   & Fake News & Real News \\

\hline
IP  & 0.504  & 0.424   & 0.557   \\
IS  & 0.827  & 0.183   & 0.903   \\
IS+IP  & 0.611  & 0.467  & 0.675   \\
T  & 0.883  & 0.872  & 0.868   \\
IP+T  & 0.811  & 0.063  & 0.896   \\
IS+T  & 0.819  & 0.128  & 0.898   \\
IS+IP+T  & 0.816  & 0.099  & 0.897   \\
\hline
${S}_\emph{m}$ reweigh. Multi-view. & 0.794  & 0.001 & 0.883\\
w/o Feature Reweigh. & 0.794  & 0.008 & 0.882\\
w/o Coarse Class. & 0.862 & 0.658 & 0.913\\
using ViT Blocks for Refine. & 0.890  & 0.689 & 0.932\\
w/o Cross. Correlat. & 0.880  & 0.589 & 0.928\\
using ViT instead of MAE & 0.884  & 0.675  & 0.927\\
w/o improving MMoE & 0.885  & 0.685  & 0.930\\
\textbf{BMR}  & \textbf{0.895}  & \textbf{0.691} & \textbf{0.936} \\
\hline
\end{tabular}}
    \caption{Ablation study on GossipCop.}
\label{table_ablation_gossip}
\end{table}

\begin{table}[!t]
\centering
  \resizebox{0.48\textwidth}{!}{
\begin{tabular}{cccc}
\hline
\multirow{2}{*}{Test} &\multirow{2}{*}{Accuracy} &\multicolumn{2}{c}{F1 Score} \\
  \cline{3-4}  &   & Fake News & Real News \\

\hline
IP  & 0.617  & 0.168   & 0.661   \\
IS  & 0.754  & 0.239   & 0.792   \\
IS+IP  & 0.770  & 0.448  & 0.842   \\
T  & 0.483  & 0.652  & 0.0   \\
IP+T  & 0.791  & 0.520  & 0.827   \\
IS+T  & 0.805  & 0.609  & 0.855   \\
IS+IP+T  & 0.882  & 0.865  & 0.894   \\
\hline
${S}_\emph{m}$ reweigh. Multi-view. & 0.875  & 0.865 & 0.860\\
w/o Feature Reweigh. & 0.892  & 0.888 & 0.884\\
w/o Coarse Class. & 0.900 & 0.895 & 0.890\\
using ViT Blocks for Refine. & 0.912  & 0.900 & 0.905\\
w/o Cross. Correlat. & 0.915  & 0.903 & 0.911\\
using ViT instead of MAE & 0.917  & 0.915  & 0.875\\
w/o improving MMoE & 0.917  & 0.914  & 0.880\\
\textbf{BMR}  & \textbf{0.929}  & \textbf{0.927} & \textbf{0.925} \\
\hline
\end{tabular}}
    \caption{Ablation study on Weibo-21.}
\label{table_ablation_weibo21}
\end{table}

\begin{figure}[!t]
  \centering
  \includegraphics[width=1.0\linewidth]{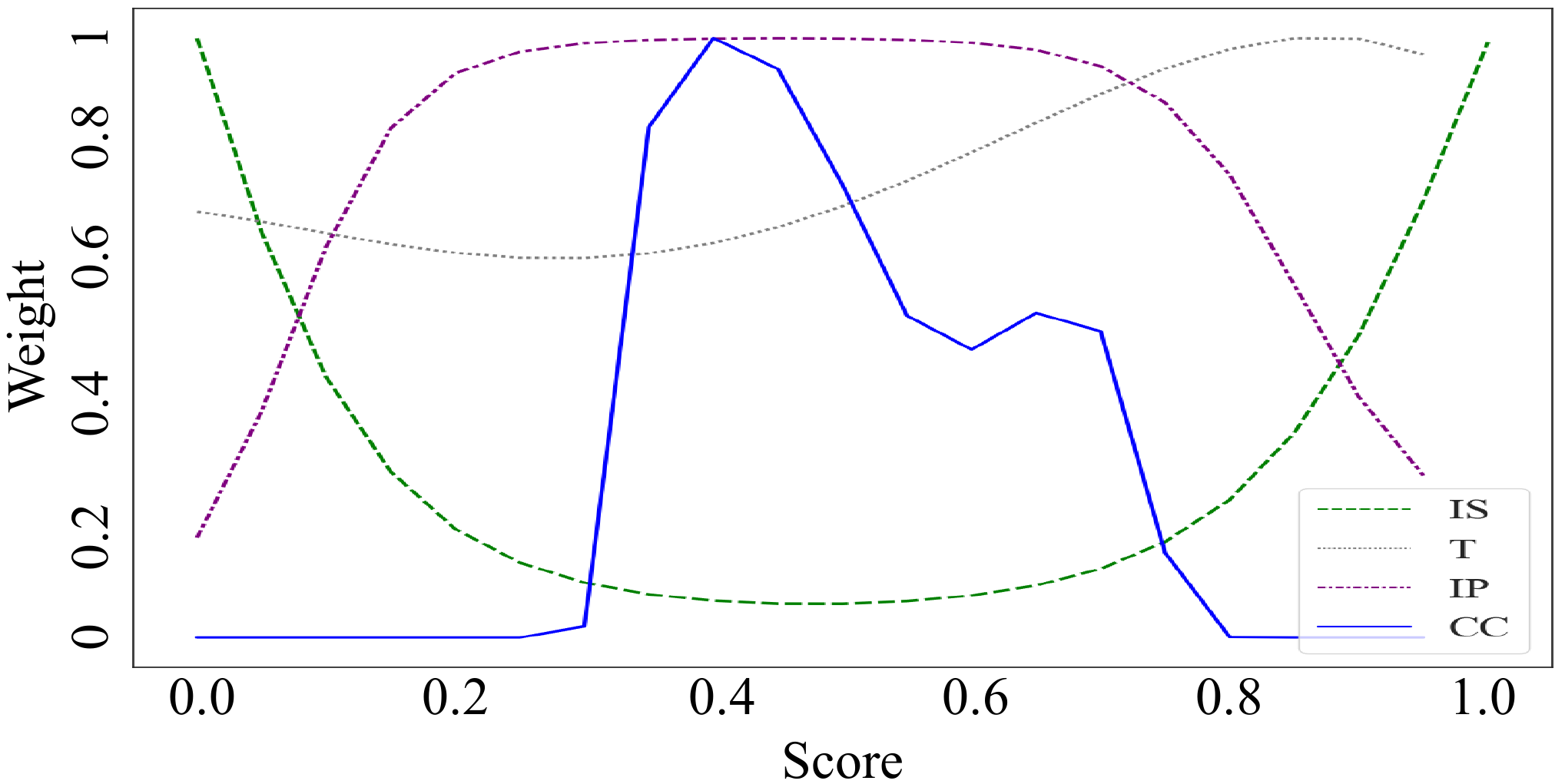}
  \caption{Curves of the learned reweighing functions $F(\cdot)$, on Weibo-21.}
%   We see little trace of overfitting and high correlation between coarse text classification and the bootstrapped result.
  \label{image_score_weibo21}
\end{figure}
\noindent\textbf{Understanding the Reweighing Curves.}
In Fig.~\ref{image_score_weibo21}, we show the reweighing scores on Weibo-21, which shows similar statistical result with those curves in Fig.8 of the main paper.
We find that $F_\emph{is}$ is U-shaped both on Weibo and Weibo-21, suggesting that if the semantics is predicted correct or wrong, not ambiguous, the corresponding representation will be bypassed. 
In contrast, $F_\emph{ip}$ tends to favor images representations which show much ambiguity in the pattern, i.e., traces that might unveil tampering or dual compression. In the real world, we can rarely assert that a news is fake if the image is modified, because we have to further distinguish legal daily image editing from malicious ones that cause misinformation. Besides, attackers usually conceal their manipulation behavior by post-processing the image, resulting in uncertainty in the prediction made by image pattern predictor.
Also, $F_\emph{cc}$ only activates the token for irrelevance when the score is approximnately within [0.3,0.8], and we believe the reason is that for both real news and fake news, the images can be either closely related or completely unrelated with the text. If there is certain consistency, there can be information that cannot be inferred from another modality, thus offering opportunity for fake news detection.

\begin{table*}[!t]
\centering
  \resizebox{1.0\textwidth}{!}{
\begin{tabular}{cccccccccccc}
\hline

   Dataset \& Label & Score & [0-0.1]  & (0.1-0.2]  & (0.2-0.3]  & (0.3-0.4]  & (0.4-0.5]  & (0.5-0.6]  & (0.6-0.7]  & (0.7-0.8]  & (0.8-0.9]  & (0.9-1.0]         \\
\hline
\multirow{5}{*}{Weibo\&Real} 

    & $\tilde{y}$   & 81.7\% & 2.0\% & 1.3\% & 1.3\% & 1.1\% & 0.3\% & 0.6\% & 1.7\% & 0.7\% & 9.1\%  \\
    & $S_{m}$   & 35.4\% & 9.8\% & 5.9\% & 4.8\% & 3.7\% & 4.6\% & 4.2\% & 5.7\% & 5.2\% & 20.6\%  \\
    & $S_{\emph{is}}$   & 53.6\% & 10.0\% & 4.7\% & 4.8\% & 3.0\% & 3.3\% & 2.7\% & 3.0\% & 3.6\% & 9.8\%  \\
    & $S_{\emph{t}}$   & 76.8\% & 3.6\% & 1.4\% & 1.8\% & 0.6\% & 1.2\% & 1.4\% & 1.7\% & 1.6\% & 9.8\%  \\
    & $S_{\emph{ip}}$   & 51.2\% & 7.6\% & 5.0\% & 3.6\% & 2.4\% & 1.7\% & 3.0\% & 3.4\% & 3.0\% & 19.1\%  \\
\hline
\multirow{5}{*}{Weibo\&Fake} 
    & $\tilde{y}$   & 3.7\% & 0.6\% & 0.7\% & 0.2\% & 0.4\% & 0.9\% & 0.3\% & 0.4\% & 1.6\% & 90.8\%  \\
    & $S_{m}$   & 14.8\% & 8.3\% & 5.0\% & 5.7\% & 3.2\% & 5.2\% & 5.0\% & 4.6\% & 8.3\% & 39.7\%  \\
    & $S_{\emph{is}}$   & 8.8\% & 4.6\% & 3.6\% & 2.9\% & 3.6\% & 4.1\% & 3.6\% & 6.0\% & 7.5\% & 55.2\%  \\
    & $S_{\emph{t}}$   & 6.3\% & 0.7\% & 0.7\% & 0.6\% & 0.7\% & 1.0\% & 0.6\% & 0.7\% & 2.5\% & 86.3\%  \\
    & $S_{\emph{ip}}$   & 8.2\% & 2.0\% & 2.0\% & 1.0\% & 1.2\% & 1.6\% & 1.5\% & 2.0\% & 3.0\% & 77.3\%  \\
\hline
\multirow{5}{*}{GossipCop\&Real} 
    & $\tilde{y}$   & 96.2\% & 0.1\% & 0\% & 0\% & 0.1\% & 0.2\% & 0.1\% & 0.1\% & 0.2\% & 3.0\%  \\
    & $S_{m}$   & 0\% & 0\% & 0.2\% & 0.5\% & 81.4\% & 17.6\% & 0.4\% & 0\% & 0\% & 0\%  \\
    & $S_{\emph{is}}$   & 49.6\% & 14.7\% & 7.6\% & 5.7\% & 2.9\% & 3.9\% & 3.0\% & 2.5\% & 3.8\% & 6.3\%  \\
    & $S_{\emph{t}}$   & 94.2\% & 0.3\% & 0.2\% & 0.2\% & 0.2\% & 0.2\% & 0.2\% & 0.2\% & 0.3\% & 3.8\%  \\
    & $S_{\emph{ip}}$   & 1.1\% & 65.4\% & 27.5\% & 5.1\% & 0.6\% & 0\% & 0.1\% & 0\% & 0\% & 0\%  \\
\hline
\multirow{5}{*}{GossipCop\&Fake} 
    & $\tilde{y}$   & 34.9\% & 0.4\% & 0.2\% & 0.4\% & 0.4\% & 0.8\% & 1.2\% & 1.2\% & 1.2\% & 59.4\%  \\
    & $S_{m}$   & 0\% & 0\% & 0.9\% & 6.2\% & 52.7\% & 33.6\% & 5.7\% & 0.9\% & 0\% & 0\%  \\
    & $S_{\emph{is}}$   & 31.7\% & 14.3\% & 8.1\% & 5.7\% & 5.7\% & 4.0\% & 4.4\% & 3.5\% & 3.3\% & 19.3\%  \\
    & $S_{\emph{t}}$   & 36.9\% & 2\% & 0\% & 0.2\% & 0.2\% & 0.2\% & 1.0\% & 0.4\% & 0.6\% & 60.3\%  \\
    & $S_{\emph{ip}}$   & 0.9\% & 51.0\% & 35.0\% & 8.1\% & 1.5\% & 0.2\% & 0\% & 0.2\% & 0\% & 3.1\%  \\
 \hline
\multirow{5}{*}{Weibo21\&Real} 
    & $\tilde{y}$   & 82.8\% & 6.5\% & 0.6\% & 0.3\% & 0.3\% & 1.0\% & 0.6\% & 0.6\% & 0.6\% & 6.5\%  \\
    & $S_{m}$   & 17.8\% & 9.7\% & 9.1\% & 11.0\% & 15.6\% & 7.5\% & 6.5\% & 5.5\% & 5.5\% & 11.7\%  \\
    & $S_{\emph{is}}$   & 24.0\% & 9.1\% & 12.7\% & 11.7\% & 10.1\% & 8.8\% & 10.4\% & 7.5\% & 4.5\% & 1.3\%  \\
    & $S_{\emph{t}}$   & 94.8\% & 5\% & 1\% & 2\% & 6\% & 2\% & 0\% & 0\% & 0\% & 0\%  \\
    & $S_{\emph{ip}}$   & 27.9\% & 8.1\% & 7.5\% & 6.5\% & 4.9\% & 4.5\% & 5.5\% & 9.4\% & 8.8\% & 16.9\%  \\
    \hline
\multirow{5}{*}{Weibo21\&Fake} 
    & $\tilde{y}$   & 3.0\% & 1.3\% & 0.7\% & 0.7\% & 0\% & 2.0\% & 2.0\% & 2.0\% & 2.3\% & 86.1\%  \\
    & $S_{m}$   & 9.2\% & 6.3\% & 8.6\% & 14.9\% & 20.46\% & 10.6\% & 8.2\% & 5.6\% & 4.6\% & 11.6\%  \\
    & $S_{\emph{is}}$   & 1.3\% & 1.3\% & 3.0\% & 4.0\% & 2.0\% & 6.6\% & 21.1\% & 13.5\% & 15.5\% & 31.7\%  \\
    & $S_{\emph{t}}$   & 14.8\% & 5.6\% & 3.6\% & 15.5\% & 45.9\% & 14.5\% & 0\% & 0\% & 0\% & 0\%  \\
    & $S_{\emph{ip}}$   & 3.3\% & 2.0\% & 1.0\% & 1.3\% & 2.0\% & 2.0\% & 2.6\% & 5.0\% & 9.2\% & 71.6\%  \\
 \hline
  \end{tabular}}
    \caption{Distribution of the scores of the ultimate classification score and the single-view prediction on three datasets.}
  \label{table_distribution_scores}
\end{table*}

\subsection{Distribution of the Predicted Scores}
In Table~\ref{table_distribution_scores}, we report the distribution of the scores of the ultimate classification score and the single-view prediction on three datasets.
Note that the results of single-view predictions are different from training BMR with these views only, because the multi-view representations in BMR are designed to be further reweighed and used by the bootstrapping module.
We surprisingly find that nearly all predictors tend to predict the two extremes, i.e., scores close to zero or one. 
The reason might be that training BMR using hard labels makes the network over-confident in the results.
Therefore, we have also tried training BMR using soft labels, i.e., for a real news, the label is changed as 0.1 rather than 0, and for a fake news, the soft label is 0.9. But the averaged performance dropped.
Besides, the single-view text predictor only predicts within range [0,0.6] on Weibo-21. Along with the fact that we failed to train a FND model on Weibo-21 using text only, we suspect that there might be more significant gaps in the statictical distribution of texts in Weibo-21 that chunk the datasets into several parts, for example, according to events, topics, etc.

\begin{figure}[!t]
  \centering
  \includegraphics[width=1.0\linewidth]{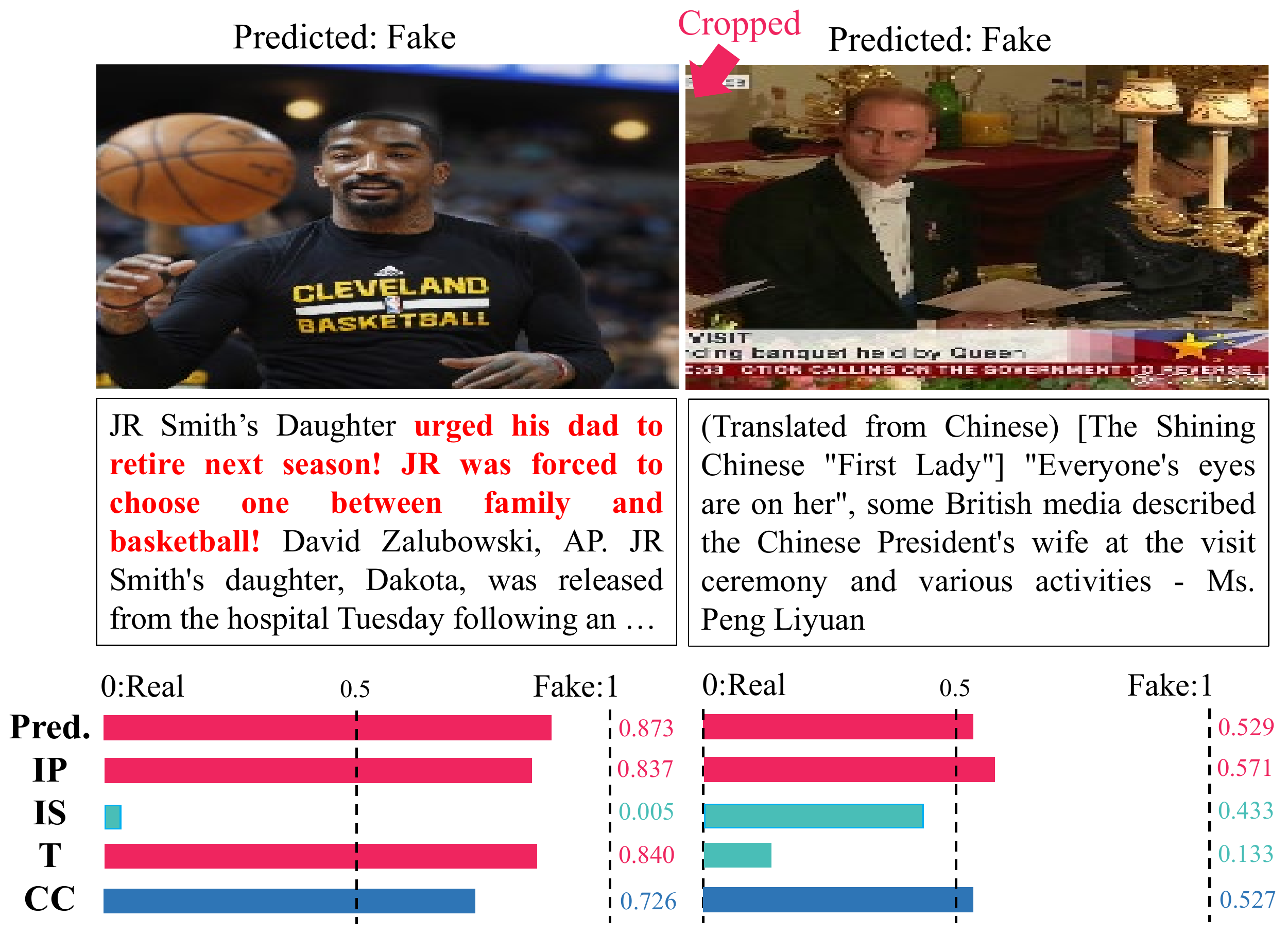}
  \caption{Sensitivity test. We distort the meaning of the two real news in Fig.1 of the main paper. The difference is marked red. The predictions are different from the original ones accordingly.}
  \label{image_modification}
\end{figure}
\subsection{Sensitivity Test}
In Fig.~\ref{image_modification}, we manually distort the two real news provided in Fig.1 of the main paper where the meaning has been altered. We invited five volunteers to do the task and to get rid of ambiguity, they must all agree that every single modified news is considered fake after modification. From the results, we see that the predictions correlated to the manipulated modality are different from the original ones in Fig.1 of the main paper, suggesting that BMR is sensitive to these differences.

We have conducted a simple quantitative experiment on the sensitivity test, in which the volunteers manipulated 50 reals news from the test set of Gossip, Weibo and Weibo-21. The manipulations are conducted under the same principle, i.e., there must be a reason why after performing such a modification, the generated news can be regarded as fake.
Then, we sent the 50 news respectively into BMR, CAFE and EANN. The results show that 37 (74\%) of these news are predicted fake by BMR, compared to 23 (46\%) of CAFE and 15 (30\%) of EANN. 
Therefore, our scheme somehow show stronger capablity of adapting to the real-world news manipulation detection circumstances. 
Note that the accuracy of the sensitivity test is lower than that on the test set of Weibo or GossipCop, because we are subjective in regarding the modified news as fake, which can be biased.

\subsection{Source Code and Data Processing}
We have attached the \textbf{anonymized} source code of BMR in the supplementary material. Because of the restriction on upload size, we are unable to upload pretrained models and the processed data. We will further open-source them on GitHub after the anonymous reviewing process.

% \bibliographystyle{aaai23}
% \bibliography{aaai23}

%% file: command.tex
\definecolor{green}{rgb}{0, 0.5, 0}
\definecolor{orange}{rgb}{0.8, 0.6, 0.2}
\definecolor{orange2}{rgb}{1.0, 0.6, 0.2}
\definecolor{red}{rgb}{1.0, 0.0, 0.0}
\definecolor{teal}{rgb}{0.0, 0.4, 0.4}
\definecolor{purple}{rgb}{0.65,0,0.65}
\definecolor{saffron}{rgb}{0.95,0.75,0.2}
\definecolor{turquoise}{rgb}{0.0,0.5,0.5}
\definecolor{black}{rgb}{0.0, 0.0, 0.0}
\definecolor{gray}{rgb}{0.5, 0.5, 0.5}
\definecolor{blue}{rgb}{0.0, 0.0, 1.0}

\newcommand{\bluemarker}[1]{{\color{blue}#1}}
\newcommand{\redmarker}[1]{{\color{red}#1}}

%% file: anonymous-submission-latex-2023.bbl
\begin{thebibliography}{93}
\providecommand{\natexlab}[1]{#1}

\bibitem[{Abdelnabi, Hasan, and Fritz(2022)}]{abdelnabi2022open}
Abdelnabi, S.; Hasan, R.; and Fritz, M. 2022.
\newblock Open-Domain, Content-based, Multi-modal Fact-checking of
  Out-of-Context Images via Online Resources.
\newblock In \emph{Proceedings of the IEEE/CVF Conference on Computer Vision
  and Pattern Recognition}, 14940--14949.

\bibitem[{Allein, Moens, and Perrotta(2021)}]{allein2021like}
Allein, L.; Moens, M.-F.; and Perrotta, D. 2021.
\newblock Like Article, Like Audience: Enforcing Multimodal Correlations for
  Disinformation Detection.
\newblock \emph{arXiv preprint arXiv:2108.13892}.

\bibitem[{Baltru{\v{s}}aitis, Ahuja, and Morency(2018)}]{Multimodal-Review}
Baltru{\v{s}}aitis, T.; Ahuja, C.; and Morency, L.-P. 2018.
\newblock Multimodal machine learning: A survey and taxonomy.
\newblock \emph{IEEE transactions on pattern analysis and machine
  intelligence}, 41(2): 423--443.

\bibitem[{Bayar and Stamm(2018)}]{bayar2018constrained}
Bayar, B.; and Stamm, M.~C. 2018.
\newblock Constrained convolutional neural networks: A new approach towards
  general purpose image manipulation detection.
\newblock \emph{IEEE Transactions on Information Forensics and Security},
  13(11): 2691--2706.

\bibitem[{Beel et~al.(2016)Beel, Gipp, Langer, and Breitinger}]{beel2016paper}
Beel, J.; Gipp, B.; Langer, S.; and Breitinger, C. 2016.
\newblock Paper recommender systems: a literature survey.
\newblock \emph{International Journal on Digital Libraries}, 17(4): 305--338.

\bibitem[{Bhatt et~al.(2018)Bhatt, Sharma, Sharma, Nagpal, Raman, and
  Mittal}]{WWW-3}
Bhatt, G.; Sharma, A.; Sharma, S.; Nagpal, A.; Raman, B.; and Mittal, A. 2018.
\newblock Combining neural, statistical and external features for fake news
  stance identification.
\newblock In \emph{Companion Proceedings of the The Web Conference 2018},
  1353--1357.

\bibitem[{Bhattarai, Granmo, and Jiao(2021)}]{TM}
Bhattarai, B.; Granmo, O.-C.; and Jiao, L. 2021.
\newblock Explainable Tsetlin Machine framework for fake news detection with
  credibility score assessment.
\newblock \emph{arXiv preprint arXiv:2105.09114}.

\bibitem[{Bian et~al.(2020)Bian, Xiao, Xu, Zhao, Huang, Rong, and
  Huang}]{WWW-4}
Bian, T.; Xiao, X.; Xu, T.; Zhao, P.; Huang, W.; Rong, Y.; and Huang, J. 2020.
\newblock Rumor detection on social media with bi-directional graph
  convolutional networks.
\newblock In \emph{Proceedings of the AAAI conference on artificial
  intelligence}, volume~34, 549--556.

\bibitem[{Boididou et~al.(2018)Boididou, Papadopoulos, Zampoglou, Apostolidis,
  Papadopoulou, and Kompatsiaris}]{Twitter}
Boididou, C.; Papadopoulos, S.; Zampoglou, M.; Apostolidis, L.; Papadopoulou,
  O.; and Kompatsiaris, Y. 2018.
\newblock Detection and visualization of misleading content on Twitter.
\newblock \emph{International Journal of Multimedia Information Retrieval},
  7(1): 71--86.

\bibitem[{Bouville(2008)}]{c:22}
Bouville, M. 2008.
\newblock Crime and punishment in scientific research.
\newblock arXiv:0803.4058.

\bibitem[{Brown et~al.(2020)Brown, Mann, Ryder, Subbiah, Kaplan, Dhariwal,
  Neelakantan, Shyam, Sastry, Askell et~al.}]{GPT}
Brown, T.; Mann, B.; Ryder, N.; Subbiah, M.; Kaplan, J.~D.; Dhariwal, P.;
  Neelakantan, A.; Shyam, P.; Sastry, G.; Askell, A.; et~al. 2020.
\newblock Language models are few-shot learners.
\newblock \emph{Advances in neural information processing systems}, 33:
  1877--1901.

\bibitem[{Cao et~al.(2020)Cao, Qi, Sheng, Yang, Guo, and Li}]{Leveraging_5}
Cao, J.; Qi, P.; Sheng, Q.; Yang, T.; Guo, J.; and Li, J. 2020.
\newblock Exploring the role of visual content in fake news detection.
\newblock \emph{Disinformation, Misinformation, and Fake News in Social Media},
  141--161.

\bibitem[{Chen et~al.(2021{\natexlab{a}})Chen, Wu, Yang, Xie, Wang, and
  Liu}]{chen2021multimodal}
Chen, J.; Wu, Z.; Yang, Z.; Xie, H.; Wang, F.~L.; and Liu, W.
  2021{\natexlab{a}}.
\newblock Multimodal fusion network with latent topic memory for rumor
  detection.
\newblock In \emph{2021 IEEE International Conference on Multimedia and Expo
  (ICME)}, 1--6. IEEE.

\bibitem[{Chen et~al.(2021{\natexlab{b}})Chen, Dong, Ji, Cao, and Li}]{MVSS}
Chen, X.; Dong, C.; Ji, J.; Cao, J.; and Li, X. 2021{\natexlab{b}}.
\newblock Image Manipulation Detection by Multi-View Multi-Scale Supervision.
\newblock In \emph{Proceedings of the IEEE/CVF International Conference on
  Computer Vision}, 14185--14193.

\bibitem[{Chen, Hu, and Sui(2019)}]{chen2019text}
Chen, Y.; Hu, L.; and Sui, J. 2019.
\newblock Text-based fusion neural network for rumor detection.
\newblock In \emph{International Conference on Knowledge Science, Engineering
  and Management}, 105--109. Springer.

\bibitem[{Chen et~al.(2022)Chen, Li, Zhang, Sui, Lv, Tun, and Shang}]{WWW}
Chen, Y.; Li, D.; Zhang, P.; Sui, J.; Lv, Q.; Tun, L.; and Shang, L. 2022.
\newblock Cross-modal Ambiguity Learning for Multimodal Fake News Detection.
\newblock In \emph{Proceedings of the ACM Web Conference 2022}, 2897--2905.

\bibitem[{Chen et~al.(2019)Chen, Sui, Hu, and Gong}]{chen2019attention}
Chen, Y.; Sui, J.; Hu, L.; and Gong, W. 2019.
\newblock Attention-residual network with CNN for rumor detection.
\newblock In \emph{Proceedings of the 28th ACM international conference on
  information and knowledge management}, 1121--1130.

\bibitem[{Child et~al.(2019)Child, Gray, Radford, and
  Sutskever}]{child2019generating}
Child, R.; Gray, S.; Radford, A.; and Sutskever, I. 2019.
\newblock Generating long sequences with sparse transformers.
\newblock \emph{arXiv preprint arXiv:1904.10509}.

\bibitem[{Choromanski et~al.(2020)Choromanski, Likhosherstov, Dohan, Song,
  Gane, Sarlos, Hawkins, Davis, Mohiuddin, Kaiser
  et~al.}]{choromanski2020rethinking}
Choromanski, K.; Likhosherstov, V.; Dohan, D.; Song, X.; Gane, A.; Sarlos, T.;
  Hawkins, P.; Davis, J.; Mohiuddin, A.; Kaiser, L.; et~al. 2020.
\newblock Rethinking attention with performers.
\newblock \emph{arXiv preprint arXiv:2009.14794}.

\bibitem[{Clancey(1979)}]{c:79}
Clancey, W.~J. 1979.
\newblock \emph{{Transfer of Rule-Based Expertise through a Tutorial
  Dialogue}}.
\newblock {Ph.D.} diss., Dept.\ of Computer Science, Stanford Univ., Stanford,
  Calif.

\bibitem[{Clancey(1983)}]{c:83}
Clancey, W.~J. 1983.
\newblock {Communication, Simulation, and Intelligent Agents: Implications of
  Personal Intelligent Machines for Medical Education}.
\newblock In \emph{Proceedings of the Eighth International Joint Conference on
  Artificial Intelligence {(IJCAI-83)}}, 556--560. Menlo Park, Calif: {IJCAI
  Organization}.

\bibitem[{Clancey(1984)}]{c:84}
Clancey, W.~J. 1984.
\newblock {Classification Problem Solving}.
\newblock In \emph{Proceedings of the Fourth National Conference on Artificial
  Intelligence}, 45--54. Menlo Park, Calif.: AAAI Press.

\bibitem[{Clancey(2021)}]{c:21}
Clancey, W.~J. 2021.
\newblock {The Engineering of Qualitative Models}.
\newblock Forthcoming.

\bibitem[{Conde and Turgutlu(2021)}]{CLIP-Art}
Conde, M.~V.; and Turgutlu, K. 2021.
\newblock CLIP-Art: contrastive pre-training for fine-grained art
  classification.
\newblock In \emph{Proceedings of the IEEE/CVF Conference on Computer Vision
  and Pattern Recognition}, 3956--3960.

\bibitem[{Conroy, Rubin, and Chen(2015)}]{Leveraging_6}
Conroy, N.~K.; Rubin, V.~L.; and Chen, Y. 2015.
\newblock Automatic deception detection: Methods for finding fake news.
\newblock \emph{Proceedings of the association for information science and
  technology}, 52(1): 1--4.

\bibitem[{Dancette et~al.(2021)Dancette, Cadene, Teney, and Cord}]{VQA}
Dancette, C.; Cadene, R.; Teney, D.; and Cord, M. 2021.
\newblock Beyond question-based biases: Assessing multimodal shortcut learning
  in visual question answering.
\newblock In \emph{Proceedings of the IEEE/CVF International Conference on
  Computer Vision}, 1574--1583.

\bibitem[{Devlin et~al.(2018{\natexlab{a}})Devlin, Chang, Lee, and
  Toutanova}]{devlin2018bert}
Devlin, J.; Chang, M.-W.; Lee, K.; and Toutanova, K. 2018{\natexlab{a}}.
\newblock Bert: Pre-training of deep bidirectional transformers for language
  understanding.
\newblock \emph{arXiv preprint arXiv:1810.04805}.

\bibitem[{Devlin et~al.(2018{\natexlab{b}})Devlin, Chang, Lee, and
  Toutanova}]{BERT}
Devlin, J.; Chang, M.-W.; Lee, K.; and Toutanova, K. 2018{\natexlab{b}}.
\newblock Bert: Pre-training of deep bidirectional transformers for language
  understanding.
\newblock \emph{arXiv preprint arXiv:1810.04805}.

\bibitem[{Engelmore and Morgan(1986)}]{em:86}
Engelmore, R.; and Morgan, A., eds. 1986.
\newblock \emph{Blackboard Systems}.
\newblock Reading, Mass.: Addison-Wesley.

\bibitem[{Gabbay, Cohen, and Hoshen(2021)}]{ViT}
Gabbay, A.; Cohen, N.; and Hoshen, Y. 2021.
\newblock An image is worth more than a thousand words: Towards disentanglement
  in the wild.
\newblock \emph{Advances in Neural Information Processing Systems}, 34.

\bibitem[{Guo et~al.(2018)Guo, Cao, Zhang, Guo, and Li}]{guo2018rumor}
Guo, H.; Cao, J.; Zhang, Y.; Guo, J.; and Li, J. 2018.
\newblock Rumor detection with hierarchical social attention network.
\newblock In \emph{Proceedings of the 27th ACM international conference on
  information and knowledge management}, 943--951.

\bibitem[{Han et~al.(2021)Han, Han, Zhang, Li, and Cao}]{han2021fighting}
Han, B.; Han, X.; Zhang, H.; Li, J.; and Cao, X. 2021.
\newblock Fighting fake news: two stream network for deepfake detection via
  learnable SRM.
\newblock \emph{IEEE Transactions on Biometrics, Behavior, and Identity
  Science}, 3(3): 320--331.

\bibitem[{Hasling, Clancey, and Rennels(1984)}]{hcr:83}
Hasling, D.~W.; Clancey, W.~J.; and Rennels, G. 1984.
\newblock Strategic explanations for a diagnostic consultation system.
\newblock \emph{International Journal of Man-Machine Studies}, 20(1): 3--19.

\bibitem[{Hasling et~al.(1983)Hasling, Clancey, Rennels, and Test}]{hcrt:83}
Hasling, D.~W.; Clancey, W.~J.; Rennels, G.~R.; and Test, T. 1983.
\newblock {Strategic Explanations in Consultation---Duplicate}.
\newblock \emph{The International Journal of Man-Machine Studies}, 20(1):
  3--19.

\bibitem[{He and Garcia(2009)}]{imbalance}
He, H.; and Garcia, E.~A. 2009.
\newblock Learning from imbalanced data.
\newblock \emph{IEEE Transactions on knowledge and data engineering}, 21(9):
  1263--1284.

\bibitem[{He et~al.(2022)He, Chen, Xie, Li, Doll{\'a}r, and
  Girshick}]{he2022masked}
He, K.; Chen, X.; Xie, S.; Li, Y.; Doll{\'a}r, P.; and Girshick, R. 2022.
\newblock Masked autoencoders are scalable vision learners.
\newblock In \emph{Proceedings of the IEEE/CVF Conference on Computer Vision
  and Pattern Recognition}, 16000--16009.

\bibitem[{He et~al.(2016{\natexlab{a}})He, Zhang, Ren, and Sun}]{he2016deep}
He, K.; Zhang, X.; Ren, S.; and Sun, J. 2016{\natexlab{a}}.
\newblock Deep residual learning for image recognition.
\newblock In \emph{Proceedings of the IEEE conference on computer vision and
  pattern recognition}, 770--778.

\bibitem[{He et~al.(2016{\natexlab{b}})He, Zhang, Ren, and Sun}]{RESNET}
He, K.; Zhang, X.; Ren, S.; and Sun, J. 2016{\natexlab{b}}.
\newblock Deep residual learning for image recognition.
\newblock In \emph{Proceedings of the IEEE conference on computer vision and
  pattern recognition}, 770--778.

\bibitem[{Hendrycks and Gimpel(2016)}]{GELU}
Hendrycks, D.; and Gimpel, K. 2016.
\newblock Gaussian error linear units (gelus).
\newblock \emph{arXiv preprint arXiv:1606.08415}.

\bibitem[{Hu, Shen, and Sun(2018)}]{hu2018squeeze}
Hu, J.; Shen, L.; and Sun, G. 2018.
\newblock Squeeze-and-excitation networks.
\newblock In \emph{Proceedings of the IEEE conference on computer vision and
  pattern recognition}, 7132--7141.

\bibitem[{Ioffe and Szegedy(2015)}]{BN}
Ioffe, S.; and Szegedy, C. 2015.
\newblock Batch normalization: Accelerating deep network training by reducing
  internal covariate shift.
\newblock In \emph{International Conference on Machine Learning (ICML)},
  448--456.

\bibitem[{Jawahar, Sagot, and Seddah(2019)}]{jawahar2019does}
Jawahar, G.; Sagot, B.; and Seddah, D. 2019.
\newblock What does BERT learn about the structure of language?
\newblock In \emph{ACL 2019-57th Annual Meeting of the Association for
  Computational Linguistics}.

\bibitem[{Jin et~al.(2017)Jin, Cao, Guo, Zhang, and Luo}]{weibo}
Jin, Z.; Cao, J.; Guo, H.; Zhang, Y.; and Luo, J. 2017.
\newblock Multimodal fusion with recurrent neural networks for rumor detection
  on microblogs.
\newblock In \emph{Proceedings of the 25th ACM international conference on
  Multimedia}, 795--816.

\bibitem[{Jin et~al.(2016)Jin, Cao, Zhang, Zhou, and Tian}]{jin2016novel}
Jin, Z.; Cao, J.; Zhang, Y.; Zhou, J.; and Tian, Q. 2016.
\newblock Novel visual and statistical image features for microblogs news
  verification.
\newblock \emph{IEEE transactions on multimedia}, 19(3): 598--608.

\bibitem[{Johnson(2012)}]{johnson2012google}
Johnson, G. 2012.
\newblock Google Translate http://translate. google. com.
\newblock \emph{Technical Services Quarterly}, 29(2): 165--165.

\bibitem[{Khattar et~al.(2019)Khattar, Goud, Gupta, and Varma}]{MVAE}
Khattar, D.; Goud, J.~S.; Gupta, M.; and Varma, V. 2019.
\newblock Mvae: Multimodal variational autoencoder for fake news detection.
\newblock In \emph{The world wide web conference}, 2915--2921.

\bibitem[{Kingma and Ba(2014)}]{kingma2014adam}
Kingma, D.~P.; and Ba, J. 2014.
\newblock Adam: A method for stochastic optimization.
\newblock \emph{arXiv preprint arXiv:1412.6980}.

\bibitem[{Li et~al.(2021)Li, Sun, Yu, Tian, Yao, and Xu}]{li2021entity}
Li, P.; Sun, X.; Yu, H.; Tian, Y.; Yao, F.; and Xu, G. 2021.
\newblock Entity-Oriented Multi-Modal Alignment and Fusion Network for Fake
  News Detection.
\newblock \emph{IEEE Transactions on Multimedia}.

\bibitem[{Li et~al.(2020)Li, Yin, Li, Zhang, Hu, Zhang, Wang, Hu, Dong, Wei
  et~al.}]{OSCAR}
Li, X.; Yin, X.; Li, C.; Zhang, P.; Hu, X.; Zhang, L.; Wang, L.; Hu, H.; Dong,
  L.; Wei, F.; et~al. 2020.
\newblock Oscar: Object-semantics aligned pre-training for vision-language
  tasks.
\newblock In \emph{European Conference on Computer Vision}, 121--137. Springer.

\bibitem[{Lin et~al.(2019)Lin, Tremblay-Taylor, Mou, You, and
  Lee}]{lin2019detecting}
Lin, J.; Tremblay-Taylor, G.; Mou, G.; You, D.; and Lee, K. 2019.
\newblock Detecting fake news articles.
\newblock In \emph{2019 IEEE International Conference on Big Data (Big Data)},
  3021--3025. IEEE.

\bibitem[{Lu et~al.(2019)Lu, Batra, Parikh, and Lee}]{vilbert}
Lu, J.; Batra, D.; Parikh, D.; and Lee, S. 2019.
\newblock Vilbert: Pretraining task-agnostic visiolinguistic representations
  for vision-and-language tasks.
\newblock \emph{Advances in neural information processing systems}, 32.

\bibitem[{Ma et~al.(2018)Ma, Zhao, Yi, Chen, Hong, and Chi}]{MMoE}
Ma, J.; Zhao, Z.; Yi, X.; Chen, J.; Hong, L.; and Chi, E.~H. 2018.
\newblock Modeling task relationships in multi-task learning with multi-gate
  mixture-of-experts.
\newblock In \emph{Proceedings of the 24th ACM SIGKDD International Conference
  on Knowledge Discovery \& Data Mining}, 1930--1939.

\bibitem[{Nan et~al.(2021)Nan, Cao, Zhu, Wang, and Li}]{MDFEND}
Nan, Q.; Cao, J.; Zhu, Y.; Wang, Y.; and Li, J. 2021.
\newblock MDFEND: Multi-domain Fake News Detection.
\newblock In \emph{Proceedings of the 30th ACM International Conference on
  Information \& Knowledge Management}, 3343--3347.

\bibitem[{{NASA}(2015)}]{c:23}
{NASA}. 2015.
\newblock Pluto: The 'Other' Red Planet.
\newblock \url{https://www.nasa.gov/nh/pluto-the-other-red-planet}.
\newblock Accessed: 2018-12-06.

\bibitem[{Nichol et~al.(2021)Nichol, Dhariwal, Ramesh, Shyam, Mishkin, McGrew,
  Sutskever, and Chen}]{Glide}
Nichol, A.; Dhariwal, P.; Ramesh, A.; Shyam, P.; Mishkin, P.; McGrew, B.;
  Sutskever, I.; and Chen, M. 2021.
\newblock Glide: Towards photorealistic image generation and editing with
  text-guided diffusion models.
\newblock \emph{arXiv preprint arXiv:2112.10741}.

\bibitem[{Ott et~al.(2018)Ott, Auli, Grangier, and Ranzato}]{ott2018analyzing}
Ott, M.; Auli, M.; Grangier, D.; and Ranzato, M. 2018.
\newblock Analyzing uncertainty in neural machine translation.
\newblock In \emph{International Conference on Machine Learning}, 3956--3965.
  PMLR.

\bibitem[{Potthast et~al.(2017)Potthast, Kiesel, Reinartz, Bevendorff, and
  Stein}]{Leveraging_27}
Potthast, M.; Kiesel, J.; Reinartz, K.; Bevendorff, J.; and Stein, B. 2017.
\newblock A stylometric inquiry into hyperpartisan and fake news.
\newblock \emph{arXiv preprint arXiv:1702.05638}.

\bibitem[{Qi et~al.(2021)Qi, Cao, Li, Liu, Sheng, Mi, He, Lv, Guo, and
  Yu}]{Entity-Enhanced}
Qi, P.; Cao, J.; Li, X.; Liu, H.; Sheng, Q.; Mi, X.; He, Q.; Lv, Y.; Guo, C.;
  and Yu, Y. 2021.
\newblock Improving Fake News Detection by Using an Entity-enhanced Framework
  to Fuse Diverse Multimodal Clues.
\newblock In \emph{Proceedings of the 29th ACM International Conference on
  Multimedia}, 1212--1220.

\bibitem[{Qi et~al.(2019{\natexlab{a}})Qi, Cao, Yang, Guo, and Li}]{ICASSP-1}
Qi, P.; Cao, J.; Yang, T.; Guo, J.; and Li, J. 2019{\natexlab{a}}.
\newblock Exploiting multi-domain visual information for fake news detection.
\newblock In \emph{2019 IEEE International Conference on Data Mining (ICDM)},
  518--527. IEEE.

\bibitem[{Qi et~al.(2019{\natexlab{b}})Qi, Cao, Yang, Guo, and
  Li}]{qi2019exploiting}
Qi, P.; Cao, J.; Yang, T.; Guo, J.; and Li, J. 2019{\natexlab{b}}.
\newblock Exploiting multi-domain visual information for fake news detection.
\newblock In \emph{2019 IEEE International Conference on Data Mining (ICDM)},
  518--527. IEEE.

\bibitem[{Qian et~al.(2018)Qian, Gong, Sharma, and Liu}]{qian2018neural}
Qian, F.; Gong, C.; Sharma, K.; and Liu, Y. 2018.
\newblock Neural User Response Generator: Fake News Detection with Collective
  User Intelligence.
\newblock In \emph{IJCAI}, volume~18, 3834--3840.

\bibitem[{Radford et~al.(2021)Radford, Kim, Hallacy, Ramesh, Goh, Agarwal,
  Sastry, Askell, Mishkin, Clark et~al.}]{CLIP}
Radford, A.; Kim, J.~W.; Hallacy, C.; Ramesh, A.; Goh, G.; Agarwal, S.; Sastry,
  G.; Askell, A.; Mishkin, P.; Clark, J.; et~al. 2021.
\newblock Learning transferable visual models from natural language
  supervision.
\newblock In \emph{International Conference on Machine Learning}, 8748--8763.
  PMLR.

\bibitem[{Raffel et~al.(2019)Raffel, Shazeer, Roberts, Lee, Narang, Matena,
  Zhou, Li, and Liu}]{raffel2019exploring}
Raffel, C.; Shazeer, N.; Roberts, A.; Lee, K.; Narang, S.; Matena, M.; Zhou,
  Y.; Li, W.; and Liu, P.~J. 2019.
\newblock Exploring the limits of transfer learning with a unified text-to-text
  transformer.
\newblock \emph{arXiv preprint arXiv:1910.10683}.

\bibitem[{Rice(1986)}]{r:86}
Rice, J. 1986.
\newblock {Poligon: A System for Parallel Problem Solving}.
\newblock Technical Report KSL-86-19, Dept.\ of Computer Science, Stanford
  Univ.

\bibitem[{Robinson(1980{\natexlab{a}})}]{r:80}
Robinson, A.~L. 1980{\natexlab{a}}.
\newblock New Ways to Make Microcircuits Smaller.
\newblock \emph{Science}, 208(4447): 1019--1022.

\bibitem[{Robinson(1980{\natexlab{b}})}]{r:80x}
Robinson, A.~L. 1980{\natexlab{b}}.
\newblock {New Ways to Make Microcircuits Smaller---Duplicate Entry}.
\newblock \emph{Science}, 208: 1019--1026.

\bibitem[{Ruchansky, Seo, and Liu(2017)}]{ruchansky2017csi}
Ruchansky, N.; Seo, S.; and Liu, Y. 2017.
\newblock Csi: A hybrid deep model for fake news detection.
\newblock In \emph{Proceedings of the 2017 ACM on Conference on Information and
  Knowledge Management}, 797--806.

\bibitem[{Shu et~al.(2020{\natexlab{a}})Shu, Mahudeswaran, Wang, Lee, and
  Liu}]{shu2020fakenewsnet}
Shu, K.; Mahudeswaran, D.; Wang, S.; Lee, D.; and Liu, H. 2020{\natexlab{a}}.
\newblock Fakenewsnet: A data repository with news content, social context, and
  spatiotemporal information for studying fake news on social media.
\newblock \emph{Big data}, 8(3): 171--188.

\bibitem[{Shu et~al.(2017)Shu, Sliva, Wang, Tang, and Liu}]{shu2017fake}
Shu, K.; Sliva, A.; Wang, S.; Tang, J.; and Liu, H. 2017.
\newblock Fake news detection on social media: A data mining perspective.
\newblock \emph{ACM SIGKDD explorations newsletter}, 19(1): 22--36.

\bibitem[{Shu et~al.(2020{\natexlab{b}})Shu, Zheng, Li, Mukherjee, Awadallah,
  Ruston, and Liu}]{shu2020leveraging}
Shu, K.; Zheng, G.; Li, Y.; Mukherjee, S.; Awadallah, A.~H.; Ruston, S.; and
  Liu, H. 2020{\natexlab{b}}.
\newblock Leveraging multi-source weak social supervision for early detection
  of fake news.
\newblock \emph{arXiv preprint arXiv:2004.01732}.

\bibitem[{Singhal et~al.(2020)Singhal, Kabra, Sharma, Shah, Chakraborty, and
  Kumaraguru}]{SpotFake}
Singhal, S.; Kabra, A.; Sharma, M.; Shah, R.~R.; Chakraborty, T.; and
  Kumaraguru, P. 2020.
\newblock Spotfake+: A multimodal framework for fake news detection via
  transfer learning (student abstract).
\newblock In \emph{Proceedings of the AAAI Conference on Artificial
  Intelligence}, volume~34, 13915--13916.

\bibitem[{Singhal et~al.(2022)Singhal, Pandey, Mrig, Shah, and
  Kumaraguru}]{singhal2022leveraging}
Singhal, S.; Pandey, T.; Mrig, S.; Shah, R.~R.; and Kumaraguru, P. 2022.
\newblock Leveraging Intra and Inter Modality Relationship for Multimodal Fake
  News Detection.

\bibitem[{Singhal et~al.(2019)Singhal, Shah, Chakraborty, Kumaraguru, and
  Satoh}]{singhal2019spotfake}
Singhal, S.; Shah, R.~R.; Chakraborty, T.; Kumaraguru, P.; and Satoh, S. 2019.
\newblock Spotfake: A multi-modal framework for fake news detection.
\newblock In \emph{2019 IEEE fifth international conference on multimedia big
  data (BigMM)}, 39--47. IEEE.

\bibitem[{Sun et~al.(2021)Sun, Zhang, Ma, and Liu}]{sun2021inconsistency}
Sun, M.; Zhang, X.; Ma, J.; and Liu, Y. 2021.
\newblock Inconsistency Matters: A Knowledge-guided Dual-inconsistency Network
  for Multi-modal Rumor Detection.
\newblock In \emph{Findings of the Association for Computational Linguistics:
  EMNLP 2021}, 1412--1423.

\bibitem[{Sun et~al.(2022)Sun, Zhang, Zheng, and Ma}]{sun2022ddgcn}
Sun, M.; Zhang, X.; Zheng, J.; and Ma, G. 2022.
\newblock DDGCN: Dual Dynamic Graph Convolutional Networks for Rumor Detection
  on Social Media.

\bibitem[{Szegedy et~al.(2016)Szegedy, Vanhoucke, Ioffe, Shlens, and
  Wojna}]{inceptionv3}
Szegedy, C.; Vanhoucke, V.; Ioffe, S.; Shlens, J.; and Wojna, Z. 2016.
\newblock Rethinking the inception architecture for computer vision.
\newblock In \emph{Proceedings of the IEEE conference on computer vision and
  pattern recognition}, 2818--2826.

\bibitem[{Van~der Maaten and Hinton(2008{\natexlab{a}})}]{van2008visualizing}
Van~der Maaten, L.; and Hinton, G. 2008{\natexlab{a}}.
\newblock Visualizing data using t-SNE.
\newblock \emph{Journal of machine learning research}, 9(11).

\bibitem[{Van~der Maaten and Hinton(2008{\natexlab{b}})}]{tsne}
Van~der Maaten, L.; and Hinton, G. 2008{\natexlab{b}}.
\newblock Visualizing data using t-SNE.
\newblock \emph{Journal of machine learning research}, 9(11).

\bibitem[{Vaswani et~al.(2017)Vaswani, Shazeer, Parmar, Uszkoreit, Jones,
  Gomez, Kaiser, and Polosukhin}]{vaswani2017attention}
Vaswani, A.; Shazeer, N.; Parmar, N.; Uszkoreit, J.; Jones, L.; Gomez, A.~N.;
  Kaiser, {\L}.; and Polosukhin, I. 2017.
\newblock Attention is all you need.
\newblock \emph{Advances in neural information processing systems}, 30.

\bibitem[{Wang et~al.(2020)Wang, Li, Khabsa, Fang, and Ma}]{wang2020linformer}
Wang, S.; Li, B.~Z.; Khabsa, M.; Fang, H.; and Ma, H. 2020.
\newblock Linformer: Self-attention with linear complexity.
\newblock \emph{arXiv preprint arXiv:2006.04768}.

\bibitem[{Wang et~al.(2018{\natexlab{a}})Wang, Girshick, Gupta, and
  He}]{wang2018non}
Wang, X.; Girshick, R.; Gupta, A.; and He, K. 2018{\natexlab{a}}.
\newblock Non-local neural networks.
\newblock In \emph{Proceedings of the IEEE conference on computer vision and
  pattern recognition}, 7794--7803.

\bibitem[{Wang et~al.(2018{\natexlab{b}})Wang, Ma, Jin, Yuan, Xun, Jha, Su, and
  Gao}]{EANN}
Wang, Y.; Ma, F.; Jin, Z.; Yuan, Y.; Xun, G.; Jha, K.; Su, L.; and Gao, J.
  2018{\natexlab{b}}.
\newblock Eann: Event adversarial neural networks for multi-modal fake news
  detection.
\newblock In \emph{Proceedings of the 24th acm sigkdd international conference
  on knowledge discovery \& data mining}, 849--857.

\bibitem[{Wei et~al.(2021)Wei, Chen, Zhou, Liao, Tan, Yuan, Zhang, and
  Yu}]{Hair-CLIP}
Wei, T.; Chen, D.; Zhou, W.; Liao, J.; Tan, Z.; Yuan, L.; Zhang, W.; and Yu, N.
  2021.
\newblock Hairclip: Design your hair by text and reference image.
\newblock \emph{arXiv preprint arXiv:2112.05142}.

\bibitem[{Wei et~al.(2022)Wei, Pan, Qiao, Niu, Dong, and Li}]{wei2022cross}
Wei, Z.; Pan, H.; Qiao, L.; Niu, X.; Dong, P.; and Li, D. 2022.
\newblock Cross-Modal Knowledge Distillation in Multi-Modal Fake News
  Detection.
\newblock In \emph{ICASSP 2022-2022 IEEE International Conference on Acoustics,
  Speech and Signal Processing (ICASSP)}, 4733--4737. IEEE.

\bibitem[{Wu et~al.(2021)Wu, Zhan, Zhang, Wang, and Xu}]{wu2021multimodal}
Wu, Y.; Zhan, P.; Zhang, Y.; Wang, L.; and Xu, Z. 2021.
\newblock Multimodal Fusion with Co-Attention Networks for Fake News Detection.
\newblock In \emph{Findings of the Association for Computational Linguistics:
  ACL-IJCNLP 2021}, 2560--2569.

\bibitem[{Xue et~al.(2021)Xue, Wang, Tian, Li, Shi, and Wei}]{xue2021detecting}
Xue, J.; Wang, Y.; Tian, Y.; Li, Y.; Shi, L.; and Wei, L. 2021.
\newblock Detecting fake news by exploring the consistency of multimodal data.
\newblock \emph{Information Processing \& Management}, 58(5): 102610.

\bibitem[{Yang et~al.(2021)Yang, Lyu, Tian, Liu, Liu, and
  Zhang}]{yang2021rumor}
Yang, X.; Lyu, Y.; Tian, T.; Liu, Y.; Liu, Y.; and Zhang, X. 2021.
\newblock Rumor detection on social media with graph structured adversarial
  learning.
\newblock In \emph{Proceedings of the twenty-ninth international conference on
  international joint conferences on artificial intelligence}, 1417--1423.

\bibitem[{Ye and Kovashka(2021)}]{VCR}
Ye, K.; and Kovashka, A. 2021.
\newblock A case study of the shortcut effects in visual commonsense reasoning.
\newblock In \emph{Proceedings of the AAAI conference on artificial
  intelligence}, volume~35, 3181--3189.

\bibitem[{Zhang et~al.(2020)Zhang, Qian, Fang, and Xu}]{zhang2020multimodal}
Zhang, H.; Qian, S.; Fang, Q.; and Xu, C. 2020.
\newblock Multimodal disentangled domain adaption for social media event rumor
  detection.
\newblock \emph{IEEE Transactions on Multimedia}, 23: 4441--4454.

\bibitem[{Zhang et~al.(2021)Zhang, Cao, Li, Sheng, Zhong, and
  Shu}]{Dual-Emotion}
Zhang, X.; Cao, J.; Li, X.; Sheng, Q.; Zhong, L.; and Shu, K. 2021.
\newblock Mining dual emotion for fake news detection.
\newblock In \emph{Proceedings of the Web Conference 2021}, 3465--3476.

\bibitem[{Zhou et~al.(2018)Zhou, Han, Morariu, and Davis}]{zhou2018learning}
Zhou, P.; Han, X.; Morariu, V.~I.; and Davis, L.~S. 2018.
\newblock Learning rich features for image manipulation detection.
\newblock In \emph{Proceedings of the IEEE Conference on Computer Vision and
  Pattern Recognition}, 1053--1061.

\bibitem[{Zhou, Wu, and Zafarani(2020)}]{zhou2020mathsf}
Zhou, X.; Wu, J.; and Zafarani, R. 2020.
\newblock SAFE: Similarity-Aware Multi-modal Fake News Detection.
\newblock In \emph{Pacific-Asia Conference on Knowledge Discovery and Data
  Mining}, 354--367. Springer.

\bibitem[{Zubiaga et~al.(2018)Zubiaga, Aker, Bontcheva, Liakata, and
  Procter}]{FND-Survey}
Zubiaga, A.; Aker, A.; Bontcheva, K.; Liakata, M.; and Procter, R. 2018.
\newblock Detection and resolution of rumours in social media: A survey.
\newblock \emph{ACM Computing Surveys (CSUR)}, 51(2): 1--36.

\end{thebibliography}
